\documentclass[twoside,11pt]{article}

\usepackage{jmlr2e}

\usepackage[utf8]{inputenc} \usepackage[T1]{fontenc}    \usepackage{amsfonts}       \usepackage{amsmath}        \usepackage{thmtools}
\usepackage{nicefrac}       \usepackage{pifont}         \usepackage{xspace}         \usepackage{subcaption}
\usepackage{bbm}

\usepackage[activate={true,nocompatibility},final,kerning=true,spacing=basictext]{microtype}      \microtypecontext{spacing=nonfrench}
\usepackage{enumitem}       \usepackage{wrapfig}        

\usepackage{hyperref}       \usepackage{url}            

\usepackage{booktabs}       \usepackage[table]{xcolor}
\usepackage{tabularx}
\usepackage{ulem}
\usepackage{multirow}
\usepackage{array}

\usepackage{graphicx}       \usepackage{tikz}
\usetikzlibrary{fit,positioning,calc,shapes,backgrounds}
\usepackage{pgfplots}
\pgfplotsset{compat=1.15}
\usepgfplotslibrary{fillbetween, groupplots}
\usepackage{caption}
\usepackage{enumitem}

\usepackage{lastpage}

\jmlrheading{24}{2023}{1-\pageref{LastPage}}{9/20; Revised
4/23}{5/23}{20-998}{Anton Tsitsulin, John Palowitch, Bryan Perozzi, and Emmanuel M\"uller}
\ShortHeadings{Graph Clustering with Graph Neural Networks}{Tsitsulin, Palowitch, Perozzi, M\"uller}

\title{Graph Clustering with Graph Neural Networks}
\author{\name Anton Tsitsulin \addr Google Research, New York, NY, USA \email{tsitsulin@google.com}  
  \AND \name John Palowitch \addr Google Research, San Francisco, CA, USA  \email{palowitch@google.com}  
  \AND \name Bryan Perozzi \addr Google Research, New York, NY, USA  \email{bperozzi@acm.org}  
  \AND \name Emmanuel M\"uller\email{emmanuel.mueller@cs.tu-dortmund.de}\\
  \addr Technical University of Dortmund, Germany
}
\editor{Honglak Lee}

\newcommand{\spara}[1]{\smallskip\noindent{\bf #1}}
\newcommand{\mpara}[1]{\medskip\noindent{\bf #1}}

\definecolor{cycle1}{RGB}{235,172,35}
\definecolor{cycle2}{RGB}{184,0,88}
\definecolor{cycle3}{RGB}{0,140,249}
\definecolor{cycle4}{RGB}{0,110,0}
\definecolor{cycle5}{RGB}{0,187,173}
\definecolor{cycle6}{RGB}{209,99,230}
\definecolor{cycle7}{RGB}{178,69,2}
\definecolor{cycle8}{RGB}{255,146,135}
\definecolor{cycle9}{RGB}{89,84,214}
\definecolor{cycle10}{RGB}{0,198,248}
\definecolor{cycle11}{RGB}{135,133,0}
\definecolor{cycle12}{RGB}{0,167,108}
\definecolor{cyclegray}{RGB}{189,189,189}

\newcommand{\cmark}{\textcolor{cycle4}{\ding{52}}} \newcommand{\xmark}{\textcolor{cycle2}{\ding{56}}}

\newcommand{\thiswork}{\textsc{DMoN}\xspace}

\newcommand{\dcora}{Cora\xspace}
\newcommand{\dciteseer}{Citeseer\xspace}
\newcommand{\dpubmed}{Pubmed\xspace}
\newcommand{\damazoncomputers}{Amazon PC\xspace}
\newcommand{\damazonphoto}{Amazon Photo\xspace}
\newcommand{\dcoauthorcs}{Coauthor CS\xspace}
\newcommand{\dcoauthorphy}{Coauthor Phys\xspace}
\newcommand{\dcoauthormed}{Coauthor Med\xspace}
\newcommand{\dcoauthorchem}{Coauthor Chem\xspace}
\newcommand{\dcoauthoreng}{Coauthor Eng\xspace}
\newcommand{\darxiv}{OGB-arXiv\xspace}

\newcommand*{\NPhard}{$\mathbf{NP}$-hard\xspace}

\newcommand*{\bigO}{\mathcal{O}}

\DeclareMathOperator{\tr}{Tr}

\newcommand*{\con}{\mathcal{C}}
\newcommand*{\modu}{\mathcal{Q}}

\newcommand*{\vD}{\mathbf{d}}
\newcommand*{\vX}{\mathbf{x}}
\newcommand*{\vY}{\mathbf{y}}

\newcommand*{\mA}{\mathbf{A}}
\newcommand*{\mB}{\mathbf{B}}
\newcommand*{\mC}{\mathbf{C}}
\newcommand*{\mD}{\mathbf{D}}
\newcommand*{\mI}{\mathbf{I}}

\newcommand*{\mP}{\mathbf{P}}

\newcommand*{\mW}{\mathbf{W}}

\newcommand*{\mX}{\mathbf{X}}

\newcommand*{\pP}{\mathbb{P}}

\newcommand*{\sR}{\mathbb{R}}

\newcommand{\ant}[1]{{#1}}
\newcommand{\antnew}[1]{{#1}}
\newcommand{\palnew}[1]{{#1}}
\newcommand{\pal}[1]{{#1}}

\begin{document}
\maketitle
\begin{abstract}Graph Neural Networks (GNNs) have achieved state-of-the-art results on many graph analysis tasks such as node classification and link prediction.
However, important \emph{unsupervised} problems on graphs, such as graph clustering, have proved more resistant to advances in GNNs.
\ant{Graph clustering has the same overall goal as node pooling in GNNs---does this mean that GNN pooling methods do a good job at clustering graphs?}

\ant{Surprisingly, the answer is no---current GNN pooling methods often fail to recover the cluster structure in cases where simple baselines, such as k-means applied on learned representations, work well.
We investigate further by carefully designing a set of experiments to study different signal-to-noise scenarios both in graph structure and attribute data.}
To address \ant{these methods' poor performance in clustering}, we introduce Deep Modularity Networks (\thiswork{}), an unsupervised pooling method inspired by the modularity measure of clustering quality, and show how it tackles recovery of the challenging clustering structure of real-world graphs.
Similarly, on real-world data, we show that \thiswork{} produces high quality clusters which correlate strongly with ground truth labels, achieving state-of-the-art results \ant{with over \textbf{40\%} improvement over other pooling methods across different metrics.}
\end{abstract}

 \section{Introduction}\label{sec:introduction}

In recent years there has been a surge of research interest in developing varieties of Graph Neural Networks (GNNs)---specialized deep learning architectures for dealing with graph-structured data, such as social networks~\citep{perozzi2014}, recommender graphs~\citep{ying2018graph}, or molecular graphs~\citep{defferrard2016}.
GNNs leverage the structure of the data as a computational graph, allowing the information to propagate across the edges of graphs~\citep{scarselli2008}.
When many real-world systems are represented as graphs, they exhibit locally inhomogeneous distributions of edges, forming \emph{clusters} (also called \emph{communities} or \emph{modules})---groups of nodes with high in-group edge density, and relatively low out-group density. Clusters can correspond to interesting phenomena in the underlying graph, for example to education \citep{traud2011comparing} or employment \citep{papacharissi2009virtual} in social graphs.
GNNs have been shown to benefit from leveraging higher-order structural information that could arise from clusters
~\citep{chen2018,ying2018}, for example through pooling or trainable attention over edges~\citep{velickovic2017}.

Interestingly, most existing work on GNNs to leverage higher-order structure does not directly address node partitioning or \palnew{cluster assignments} within the computational graph. Furthermore, most works explore these mechanisms only within a semi-supervised or supervised framework, ignoring the fact that \textit{unsupervised} graph clustering is often an extremely useful end-goal in itself---whether for data exploration \citep{perozzi2018discovering},  visualization \citep{clemenccon2012hierarchical, cui2008geometry}, genomic feature discovery \citep{cabreros2016detecting}, anomaly detection \citep{perozzi2016scalable}, or for many other use-cases discussed e.g.\ in \cite{fortunato2016community}. Additionally, many of the existing \ant{unsupervised} structure-aware methods have undesirable properties, such as relying on a multi-step optimization process which does not allow for an end-to-end differentiable objective~\citep{perozzi2014focused}.

In this work, we take an \textit{ab initio} approach to the clustering problem in the GNN domain, bridging the gap between traditional graph clustering objectives and deep neural networks.
We start by drawing a connection between graph pooling, which was typically studied in the literature as a regularizer for supervised GNN architectures, and fully unsupervised clustering.
Specifically, we contribute:

\begin{itemize}[leftmargin=0.2cm,itemindent=.3cm,labelwidth=\itemindent,labelsep=0cm,align=left,topsep=0pt,itemsep=-1ex,partopsep=0ex,parsep=1ex]
    \item \thiswork, \pal{an unsupervised clustering module for GNNs} that allows optimization of cluster assignments in an end-to-end differentiable way \ant{with strong empirical performance}.
    \item An empirical study of performance on synthetic graphs, illustrating the problems with existing work and how \thiswork allows for improved model performance in those regimes.
    \item Thorough experimental evaluation on real-world data, showing that many pooling methods poorly reflect hierarchical structures and are not able to make use of either graph structure and node attributes \ant{nor leverage joint information}.
\end{itemize} \section{Related Work}\label{sec:related-work}

\begin{table}[tb]
\small
\centering{
\newcolumntype{R}{>{\raggedleft\arraybackslash}X}
\newcolumntype{C}{>{\centering\arraybackslash}X}
\newcolumntype{L}{>{\hsize=0.75\hsize}C}
\newcolumntype{S}{>{\hsize=0.55\hsize}C}
\caption{Related work in terms of six desirable clustering properties outlined in Section~\ref{sec:related-work}.}
\label{tbl:related-work}
\vspace*{-3mm}
\begin{tabularx}{\linewidth}{p{0.9cm}LSLSLSC@{}}
\toprule
\multicolumn{2}{C}{} & \multicolumn{2}{c}{\cmark \emph{required for clustering}} \\
\emph{method} & \mbox{End-to-end} & Unsup.& \mbox{Node pooling} & Sparse & \mbox{Soft assign.} & Stable & Complexity\\
\midrule
Graclus & \xmark & \cmark & \cmark & \cmark & \xmark & \cmark & $\mathcal{O}(dn+m)$ \\
DiffPool & \cmark & \cmark & \cmark & \xmark & \cmark & \xmark & $\mathcal{O}(dn^2)$ \\
AGC & \xmark & \cmark & \cmark & \xmark & \xmark & \xmark & \mbox{$\mathcal{O}(dn^2k)$} \\
DAEGC & \xmark & \cmark & \cmark & \xmark & \xmark & \xmark & $\mathcal{O}(dn^k)$ \\
SDCN & \xmark & \cmark & \cmark & \cmark & \xmark & \xmark & $\mathcal{O}(d^2n+m)$ \\
NOCD & \cmark & \cmark & \cmark & \xmark & \cmark & \cmark & $\mathcal{O}(dn+m)$ \\
\mbox{Top-k} & \cmark & \xmark & \xmark & \cmark & \xmark & \cmark & $\mathcal{O}(dn+m)$ \\
\mbox{SAG} & \cmark & \xmark & \xmark & \cmark & \xmark & \xmark & $\mathcal{O}(dn+m)$ \\
\mbox{MinCut} & \cmark & \cmark & \cmark & \cmark & \cmark & \xmark & $\mathcal{O}(d^2n+m)$\\
\midrule
\mbox{\thiswork} & \cmark & \cmark & \cmark & \cmark & \cmark & \cmark & \mbox{$\mathcal{O}(d^2n+m)$} \\
\bottomrule
\end{tabularx}}
\end{table}
\vspace{-3mm} 
We build upon a rich line of research on graph neural networks and graph pooling methods. 

\spara{Graph Neural Networks} (GNNs)~\citep{scarselli2008,duvenaud2015,niepert2016,gilmer2017,kipf2017} allow end-to-end differentiable losses over data with arbitrary structure.  They have been applied to an incredible range of applications, from social networks \citep{perozzi2014}, to recommender systems \citep{ying2018graph}, to computational chemistry~\citep{gilmer2017}.  
While GNNs are flexible enough to allow for unsupervised losses, most work follows the semi-supervised setting for node classification from \citep{kipf2017}.
For an introduction to the vast topic we refer to detailed surveys~\citep{bronstein2017,hamilton2017survey,chami2020}.

\spara{Unsupervised training of GNNs} is commonly done via maximizing mutual information~\citep{belghazi2018,hjelm2018,tschannen2019,song2019} in a self-supervised fashion.
Deep Graph Infomax (DGI)~\citep{velickovic2018} adapted the mutual information-based learning from Deep InfoMax~\citep{hjelm2018}, learning unsupervised representations for nodes in attributed graphs.
InfoGraph~\citep{sun2020} extended the idea to learning representations of whole graphs instead of nodes. A very similar approach was independently proposed by~\citet{hu2020} in the context of pre-training GNNs for producing graph representations, which was first studied in~\citet{navarin2018}.

\spara{Graph pooling} aims to tackle the hierarchical nature of graphs via iterative coarsening.
Early architectures~\citep{defferrard2016} resorted to fixed axiomatic pooling, with no optimization of clustering while the network learns.
DiffPool~\citep{ying2018} suggests to include a \textit{learnable} pooling to GNN architecture. To help the convergence, DiffPool includes a link prediction loss to help encapsulate the clustering structure of graphs and an additional entropy loss 
to penalize soft assignments. Top-k~\citep{gao2019} and SAG pooling~\citep{lee2019} learn to sparsify the graph (select top-k edges for each node) with learned weights. MinCutPool~\citep{bianchi2020} pooling studies a differentiable formulation of spectral clustering as a pooling strategy.
\ant{As we show in Section~\ref{ssec:clustersize}, MinCutPool does not optimize its spectral objective, it merely orthogonalizes its features.}

We summarize mainstream graph pooling methods in Table~\ref{tbl:related-work} in terms of seven desirable properties related to their clustering capabilities:

\begin{itemize}[leftmargin=0.2cm,itemindent=.3cm,labelwidth=\itemindent,labelsep=0cm,align=left,topsep=0pt,itemsep=-1ex,partopsep=0ex,parsep=1ex]
    \item \textbf{End-to-end training} allows to capture both graph structure and node features. All of AGC~\citep{zhang2019attributed}, DAEGC~\citep{wang2019attributed}, and SDCN~\citep{bo2020structural} use k-means to initialize the model, prohibiting end-to-end learning.
    \item \textbf{Unsupervised} training is a desirable setting for clustering models. Works on \textit{supervised} graph clustering~\citep{yang2019,wang2019cvpr} are outside of our scope.
    \item \textbf{Node aggregation} is crucial for our interpretation of graph pooling in terms of clustering. Both Top-k and SAG pooling only sparsify the graph and do not reduce the nodeset.
    \item \textbf{Sparse.} As graphs in the real-world vary in size and sparsity, methods cannot be limited by an $\bigO(n^2)$ link prediction objectives, like DiffPool~\citep{ying2018}, or computing $\mA^t$, like top-k pooling methods~\citep{gao2019,lee2019} or AGC and DAEGC. The gradients of NOCD~\citep{shchur2019overlapping} are quadratic in the number of nodes; however, subsampling is solved to attain desirable subquadratic scalability.
    \item \textbf{Soft assignments} allow for more flexible reasoning about the interactions of clusters.
    \item \textbf{Stable} -- the method should be stable in terms of the graph structure. DiffPool, AGC, DAEGC, and SDCN change their performance significantly as the graph becomes more sparse, and MinCutPool \ant{fails to converge on power-law graphs.}
\end{itemize}

\antnew{We additionally analyze the computational complexity of the methods. Generally, non-sparse methods, for instance DiffPool, AGC, and DAEGC can not scale to large graphs, since their complexity is at least quadratic. Some methods, like NOCD, top-k pooling, and SAG subsample the quadratic terms in their objectives/optimizations to attain desirable scalability. We analyze the complexity of \thiswork in Section~\ref{ssec:dmon}. \thiswork maintains best-in-class scalability without sacrificing any information to subsampling.}

\spara{Graph
embeddings}~\citep{perozzi2014,grover2016,tsitsulin2018} can be thought of as (very restricted) unsupervised GNNs with an identity feature matrix, meaning each node learns its own positional representation~\citep{you2019}. The learning process in graph embeddings is often done in a similar way to DGI through noise contrastive estimation~\citep{gutmann2010}.
As far as we know, all pooling strategies for learning node embeddings without attributes have been rigid~\citep{chen2018,liang2018,deng2020};
\ant{making learning-based ones an interesting future research direction}. %
 \section{Preliminaries}\label{sec:preliminaries}

We introduce the necessary background for \thiswork{}, starting with the problem formulation, reviewing common graph clustering objectives and how they can be made differentiable.

A graph $G=(V,E)$ is defined via a set of nodes $V=(v_1,\ldots,v_n), |V|=n$ and edges $ E\subseteq V\times V, |E|=m$.
We denote the $n\times n$ adjacency matrix of $G$ by $\mA$, where $\mA_{ij} = 1$ iff $\{v_i, v_j\}\in E$ (otherwise entries of $\mA$ are equal to 0). The ``degree" of $v_i$ is its number of connections $d_i:=\sum_{j=1}^n\mA_{ij}$. We are interested in measuring the quality of graph partitioning function $\mathcal{F} : V\mapsto \{1, \ldots, k\}$ that splits the set of nodes $V$ into $k$ partitions $V_i=\{v_j: \mathcal{F}(v_j)=i\}$.
\pal{In contrast to standard graph clustering, we are also provided with node attributes $\mX\in\sR^{n\times s}$.
}

\subsection{Graph Clustering Quality Functions}\label{ssec:modularity}

\pal{As classical clustering objectives are discrete and therefore unsuitable for gradient-based optimization, \thiswork{} and the few prior works depend on spectral approximations.
To contextualize and motivate our contributions, we review two families} of clustering quality functions amenable to spectral optimization, and review some of their shortcomings.

\spara{Cut-based metrics.} In his seminal work, \citet{fiedler1973algebraic} suggested that the second eigenvector of a graph Laplacian produces a graph \textit{cut} minimal in terms of the weight of the edges.
This plain notion of cut degenerates on real-world graphs, as it does not require partitions to be balanced in terms of size.
It is possible to get normalized partitions with the use of ratio cut~\citep{wei1989towards}, which normalizes the cut by the product of the number of nodes in two partitions, or normalized cut~\citep{shi2000normalized}, which uses total edge volume of the partition as normalization. 

In real networks, however, there is evidence \textit{against} existence of good cuts~\citep{leskovec2008statistical} in ground-truth communities.
This can be explained by the fact that a single node implicitly participates in many different clusters~\citep{epasto2017ego}, e.g.\ a person in a social network is simultaneously connected with family and work friends, forcing the algorithm to merge these communities together.

Recently, MinCutPool~\citep{bianchi2020} adapted the notion of the normalized cut to use as a regularizer for pooling.
While MinCutPool's objective should, theoretically, be suitable for clustering nodes in graphs, we \pal{show that MinCutPool does not optimize its own objective function, using synthetic and real-world experiments.}

\spara{Modularity}. The modularity~\citep{newman2006} objective approaches the same problem from a statistical perspective, incorporating a \textit{null model} to quantify \pal{the deviation of the clustering from what would be observed in expectation under a random graph.}
In a fully random graph with given degrees, nodes $u$ and $v$ with degrees $d_u$ and $d_v$ are connected with probability $\nicefrac{d_ud_v}{2m}$.
Modularity measures the divergence between the intra-cluster edges from the expected one:
\begin{equation}\label{eq:modularity}
    \modu = \frac{1}{2m}\sum_{ij}\left[\mA_{ij}-\frac{d_id_j}{2m}\right]\delta(c_i, c_j),
\end{equation}
\noindent where $\delta(c_i, c_j)=1$ if $i$ and $j$ are in the same cluster and 0 otherwise.
Note $\modu\in(\nicefrac{-1}{2}; 1]$ (it is 0 when there is no correlation of clusters with edge density), but it is not necessarily maximized at 1, and is only comparable across graphs with the same degree distribution.
Moreover, optimal modularity is positive even for Erd\H{o}s--R\'enyi random graphs~\citep{mcdiarmid2020modularity}, meaning positive modularity does not imply strong clustering structure.
While problems with the modularity metric have been identified \citep{good2010performance}, it remains one of the most commonly-used and eminently useful graph clustering metrics in scientific literature~\citep{fortunato2016community}.

\subsection{Spectral Modularity Maximization}\label{ssec:spectral-modularity}
Maximizing the modularity is proven to be \NPhard~\citep{brandes2006maximizing}, however, a spectral relaxation of the problem can be solved efficiently~\citep{newman2006finding}.
Let $\mC\in\{0, 1\}^{n\times k}$ be the cluster assignment matrix and $\vD$ be the degree vector.
Then, with \textit{modularity matrix} $\mB$ defined as $\mB=\mA-\frac{\vD\vD^\top}{2m}$, the modularity $\modu$ can be reformulated as:
\begin{equation}\label{eq:spectral_modularity}
    \modu = \frac{1}{2m}\tr(\mC^\top\mB\mC)
\end{equation}
Relaxing $\mC\in\sR^{n\times k}$, the optimal $\mC$ maximizing $\modu$ is the top-k eigenvectors of the modularity matrix $\mB$.
While $\mB$ is dense, iterative eigenvalue solvers can take advantage of the fact that $\mB$ is a sum of a sparse $\mA$ and rank-one matrix $-\frac{\vD\vD^\top}{2m}$, meaning that the matrix-vector product $\mB\vX$ can be computed efficiently as
\begin{equation*}\label{eq:modularity-matvec}
    \mB\vX = \mA\vX - \frac{\vD^\top\vX\vD}{2m}
\end{equation*}

\noindent and optimized efficiently with iterative methods such as power iteration or Lanczos algorithm.
One can then obtain clusters by means of spectral bisection~\citep{newman2006finding} with iterative refinement akin to Kernighan-Lin algorithm~\citep{kernighan1970efficient}.
However, these formulations operate entirely on the graph structure, and it is non-trivial to adapt them to work with attributed graphs.

\subsection{Graph Neural Networks}\label{sec:gnns}

Graph Neural Networks are a flexible class of models that perform nonlinear feature aggregation with respect to graph structure.
For the purposes of this work, we consider transductive GNNs that output a single embedding per node.
Graph convolutional networks (GCNs)~\cite{kipf2017} are simple yet effective~\cite{shchur2018} message-passing networks that fit our criteria.
Let $\mX^0\in\sR^{n\times s}$ be the initial node features and $\Tilde{\mA} = \mD^{-\frac{1}{2}}\mA\mD^{-\frac{1}{2}}$ be the normalized adjacency matrix, the output of $t$-th layer $\mX^{t+1}$ is
\begin{equation}\label{eq:gcn}
    \mX^{t+1} = \mathrm{SeLU}(\Tilde{\mA}\mX^{t}\mW + \mX\mW_{\mathrm{skip}})
\end{equation}

We make two changes to the classic GCN architecture: first, we remove the self-loop creation and instead use an $\mW_{\mathrm{skip}}\in\sR^{s\times s}$ trainable skip connection, and, second, we replace ReLU nonlinearity with SeLU~\citep{klambauer2017self} for better convergence. \section{Method}\label{sec:method}

In this section, we present \thiswork, our method for attributed graph clustering with graph neural networks.
Inspired by the modularity quality function and its spectral optimization, we propose a fully differentiable unsupervised clustering objective which optimizes soft cluster assignments using a null model to control for inhomogeneities in the graph.
We then discuss the challenge of regularizing cluster assignments, and present collapse regularization that is effective at preventing trivial solutions \ant{without compromising optimization of the objective}.

\subsection{\thiswork: Deep Modularity Networks}\label{ssec:dmon}
\palnew{Our approach to GNN modularity is comprised of (1) the architecture to encode the cluster assignments $\mC$, and (2) the objective function with which to optimize the assignments. We} propose to obtain $\mC$ via the output of a softmax function, which allows the (soft) cluster assignment to be differentiable.
The input to the cluster assignment can be any differentiable message passing function, but here we specifically consider the case where a graph convolutional network is used to obtain soft clusters for each node as follows:
\begin{equation}\label{eq:dmon-architecture}
    \mC = \mathrm{softmax}(\mathrm{GCN}(\Tilde{\mA}, \mX)),
\end{equation}

\noindent where $\mathrm{GCN}$ is a (possibly) multi-layer convolutional network operating on a normalized adjacency matrix $\Tilde{\mA} = \mD^{-\frac{1}{2}}(\mA)\mD^{-\frac{1}{2}}$.

We then propose to optimize this assignment with the following objective, which combines insights from spectral modularity maximization (recall Eq.\ \eqref{eq:spectral_modularity}) with a novel regularization to prevent trivial solutions to the optimization problem:
\begin{equation}\label{eq:dmon-objective}
    \mathcal{L}_\text{{\thiswork}}(\mC; \mA) = \underbrace{-\frac{1}{2m}\tr ( \mC^\top \mB \mC)}_{\mathrm{modularity}} + 
    \underbrace{\frac{\sqrt{k}}{n}\left\lVert \sum_i\mC^\top_i\right\rVert_F - 1}_{\mathrm{collapse~regularization}}, 
\end{equation}
\noindent where $\lVert\cdot\rVert_F$ is the Frobenius norm.
We decompose the computation of $\tr(\mC^\top\mB\mC)$ as a sum of sparse matrix-matrix multiplication and rank-one degree normalization $\tr(\mC^\top\mA\mC-\mC^\top\vD^\top\vD\mC)$.
\antnew{This brings the time complexity of computing $\mathcal{L}_\text{{\thiswork}}$ down from $\mathcal{O}(n^2)$ to $\mathcal{O}(d^2n)$ per update, allowing \thiswork to efficiently work with sparse graphs.}

\subsection{Collapse regularization}\label{ssec:clustersize}

Without additional constraints on the assignment matrix $\mC$, spectral clustering for both min-cut and modularity objectives has spurious local minima: assigning all nodes to the same cluster produces a trivial locally optimal solution that traps gradient-based optimization methods.
MinCutPool addresses this problem by adapting spectral orthogonality constraint in the form of soft-orthogonality regularization $\left\lVert \mC^\top\mC - \mI\right\rVert_F$ from~\citet{bansal2018can}.

\begin{wrapfigure}[18]{R}{0.35\textwidth}
\begin{tikzpicture}
\begin{groupplot}[group style={
                      group name=myplot,
                      group size= 1 by 2,vertical sep=0.75cm},
                      height=3cm,
                      width=1.15\linewidth,
                      title style={at={(0.5,0.9)},anchor=south},
                      every axis x label/.style={at={(axis description cs:0.5,-0.2)},anchor=north},
                      yticklabels={,,},
                      xmin=0,
                      xmax=200,
                      ylabel=loss,]
\nextgroupplot[
 	title = \textbf{\thiswork{}},
 	legend columns=2,
	legend style={at={(1,1.4)},anchor=south east},
    legend entries={Objective, Regularizer},
	xticklabels={,,},
]
\addplot[very thick,color=cycle2] table[x=epoch,y=mgcn] {data/loss_histories.tex};
\addplot[very thick,color=cycle3] table[x=epoch,y=mgcn-reg] {data/loss_histories.tex};
\nextgroupplot[
 	title = \textbf{MinCutPool},
	xlabel=epoch,
]
\draw [color=black, dashed, very thick, draw opacity=0.75] (0, -0.9972) -- (200,-0.9972);
\draw[->,thick,color=black](axis cs:100,-0.2)--(axis cs:115,-0.8);
\node[anchor=east] (text) at (axis cs:100,-0.2){\tiny \textbf{not improving}};
\addplot[very thick,color=cycle2] table[x=epoch,y=mincut] {data/loss_histories.tex};
\addplot[very thick,color=cycle3] table[x=epoch,y=mincut-reg] {data/loss_histories.tex};
\end{groupplot}
\end{tikzpicture}
\vspace*{-7.5mm}
\caption{\label{fig:loss-evolution}\small Optimization progress of {MinCut} and \thiswork{} on Cora dataset. MinCut optimizes the regularizer, while \thiswork{} minimizes its main objective. Dashed line shows the initial loss value.}
\end{wrapfigure}
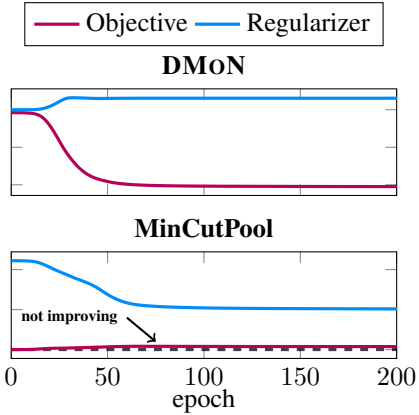 
We find that, interestingly, MinCutPool's orthogonality constraint is overly restrictive when combined with softmax class assignment.
This can be seen empirically by tracking the progress of MinCutPool optimization, shown in Figure~\ref{fig:loss-evolution}.
We observe that after 200 epochs on the CORA dataset, MinCutPool's soft orthogonality regularization term dominates its clustering term, which becomes \textit{worse than random} after training. 
We hypothesize that orthogonalizing softmax representations may be a difficult regularization objective to optimize.

Another effect of MinCutPool's harsh orthogonality constraint is seen in Table~\ref{tbl:mincut-ortho}.
We show that a constraint-only version (termed "Ortho") that we implemented actually performs \emph{better} on the downstream classification task than the original loss on all datasets except PubMed, while performing uniformly worse with respect to the graph metrics -- as expected, since it ignores the clustering loss.
This suggests that the constraint is not well-designed for the task, and may be trapping the method in local minima.
We further corroborate this in Section~\ref{ssec:real-experiments}.

We propose \textit{collapse regularization}, a more relaxed constraint, that prevents the trivial partition while not dominating the optimization of the main objective.
The regularizer is a Frobenius norm of the (soft) cluster membership counts, normalized to range $[0, \sqrt{k}]$.
It gets value of $0$ when cluster sizes are perfectly balanced, and $\sqrt{k}$ in the case all clusters collapse to one.
We also \ant{improve} the training by applying dropout~\citep{srivastava2014dropout} to GNN representations before the softmax, preventing the gradient descent from getting stuck in the local optima of the highly non-convex objective function.

\begin{table}[!t]
\footnotesize
\centering{
\newcolumntype{R}{>{\raggedleft\arraybackslash}X}
\newcolumntype{C}{>{\centering\arraybackslash}X}
\newcolumntype{S}{>{\centering\arraybackslash\hsize=.8\hsize}X}
\caption{Comparison of MinCutPool with using \textbf{only} its orthogonality regularization in terms of graph conductance $\con$, modularity $\modu$, and NMI with ground-truth labels.}
\label{tbl:mincut-ortho}
\begin{tabularx}{\linewidth}{@{}p{1.1cm}SSCSSCSSCSSCSSC}
\toprule
\multicolumn{1}{C}{} & \multicolumn{3}{c}{\textbf{\dcora}} & \multicolumn{3}{c}{\textbf{\dciteseer}} & \multicolumn{3}{c}{\textbf{\dpubmed}} & \multicolumn{3}{c}{\textbf{\dcoauthorcs}} & \multicolumn{3}{c}{\textbf{\dcoauthorphy}} \\
\multicolumn{1}{C}{} & \multicolumn{2}{c}{\emph{graph}} & \emph{labels} & \multicolumn{2}{c}{\emph{graph}} & \emph{labels} & \multicolumn{2}{c}{\emph{graph}} & \emph{labels} & \multicolumn{2}{c}{\emph{graph}} & \emph{labels} & \multicolumn{2}{c}{\emph{graph}} & \emph{labels}  \\
\cmidrule(lr){2-4}\cmidrule(lr){5-7}\cmidrule(lr){8-10}\cmidrule(lr){11-13}\cmidrule{14-16}
\emph{method} & $\con\downarrow$ & $\modu\uparrow$ & NMI$\uparrow$ & $\con\downarrow$ & $\modu\uparrow$ & NMI$\uparrow$ & $\con\downarrow$ & $\modu\uparrow$ & NMI$\uparrow$ & $\con\downarrow$ & $\modu\uparrow$ & NMI$\uparrow$ & $\con\downarrow$ & $\modu\uparrow$ & NMI$\uparrow$\\
\midrule
\mbox{MinCut} & 23.3 & 70.3 & 35.8 & 14.1 & 78.9 & 25.9 & 29.6 & 63.1 & 25.4 & 22.7 & 70.5 & 64.6 & 27.8 & 64.3 & 48.3 \\
\mbox{Ortho} & 28.0 & 65.6 & \textbf{38.4} & 18.4 & 74.5 & \textbf{26.1} & 57.8 & 32.9 & 20.3 & 27.8 & 65.7 & \textbf{64.6} & 33.0 & 59.5 & 44.7 \\
\bottomrule
\end{tabularx}}
\end{table} 
\subsection{Theoretical insights}\label{ssec:theory}

In this section we examine the validity of our proposed objective function, and in particular our novel collapse constraint. Specifically, we show two theorems:

\begin{enumerate}
    \item We prove that optimizing our collapse constraint avoids the solution with trivial clusters, and hence accomplishes the effect for which it was designed.
    \item Since the collapse constraint inherently discourages unbalanced cluster assignments (to avoid the trivial clustering), we also show that this discouragement does not affect the asymptotic performance of our overall objective function.
\end{enumerate}

The first theorem essentially tells us that that the collapse regularization can not hurt optimization in non-trivial graphs: 

\begin{theorem}
If the clustering solution encoded by the $C$ matrix has positive modularity, then our objective score $\mathcal{L}_\text{{\thiswork}}$ is smaller than the one produced by the trivial clustering.
\end{theorem}
\begin{proof}
We show that both components of the loss are smaller.
Since modularity of the trivial clustering is 0, any positive modularity will result in a smaller loss. 
Maximum of $L_2$ norm of a vector with given $L_1$ norm is achieved when only one element is nonzero -- in case of the collapse regularization, when all nodes are in one cluster. Therefore, any other partition produces a smaller value.
\end{proof}

Our second theorem is based on previous work on the ``consistency'' of graph clustering algorithms under an appropriate graph generative model. In \cite{bickel2009nonparametric} and \cite{zhao2012consistency}, a graph clustering algorithm is consistent if the global maximum clustering solution from its objective function has (in-probability) vanishing error rate in the limit of large graphs, when graphs are generated from the cluster-laden Degree-Corrected Stochastic Block Model (DC-SBM). We now detail this model and specific notions of consistency, and show that our novel, GNN-compatible objective function has consistency properties that are equivalent to standard objective functions.

The standard (degree-free) Stochastic Block Model \citep{nowicki2001estimation} is a generative graph model which divides $n$ vertices into $k$ classes, and then places edges between two vertices $v_i$ and $v_j$ with probability $p_{ij}$ determined from the class assignments. Specifically, each vertex $v_i$ is given a class $y_i\in\{1,\ldots, k\}$, and an edge $\{v_i, v_j\}$ is added to the edge set $E$ with probability $P_{y_iy_j}$, where $P$ is a symmetric $k\times k$ matrix containing the between/within-community edge probabilities. In other words, each entry $\mA_{ij}$ of the adjacency matrix is an independent Bernoulli random variable with probability $P_{y_iy_j} = \mathbb{E}[\mA_{ij}]$. Assortative clustering structure in a graph can be induced using the SBM by setting the on-diagonal probabilities of $P$ higher than the off-diagonal probabilities. Thus, the SBM has been used to analyze graph clustering algorithms both theoretically \citep{bickel2009nonparametric, abbe2015community} and empirically \citep{mothe2017community}.

The Degree-Corrected SBM \citep{karrer2011stochastic} was introduced to handle the heterogeneity of real-world degree distributions. In the DC-SBM, each node $v_i$ is given a degree parameter $\theta_i$, and edges are generated via the formula $\mathbb{E}[\mA_{ij}] = \theta_i\theta_jP_{y_iy_j}$. Thus two nodes with higher degree parameters will connect more frequently under this model.

In this paper, we adopt the theoretical setting from \cite{zhao2012consistency}: in a DC-SBM $\pP_n$ on $n$ nodes with $k$ classes, each node $v_i$ is given a label/degree pair $(y_i, \theta_i)$ drawn from a discrete joint distribution $\Pi_{k\times m}$ which is fixed and does not depend on $n$. This implies that each $\theta_i$ is one of a fixed set of values $0\leq x_1\leq\ldots\leq x_m$. To facilitate analysis of asymptotic graph sparsity, we parameterize the edge probability matrix $P$ as $P_n=\rho_nP$ where $P$ is independent of $n$, and $\rho_n = \lambda_n/n$ where $\lambda_n$ is the average degree of the network. We assume that $x_m^{2}\max P_n\leq 1.0$ so that all edge probabilities are proper. We define the marginal distribution of the labels $y$ as $\tilde{\pi} = (\tilde{\pi}_1, \ldots, \tilde{\pi}_k)$.

Given a DC-SBM $\pP_n$ on $n$ nodes with $k$ classes, we denote $\hat{\vY}_n = [\hat{y}_1, \hat{y}_2, \ldots, \hat{y}_n]$ to be a predicted class label vector, and $\hat{\mC}_n$ to be the associated $n\times k$ row-wise one-hot matrix. With $\mA^{(n)}$ the random adjacency matrix of the DC-SBM, define the random variable $\hat{\mC}^\star_n$ to be the global maximum of $\mathcal{L}_{\thiswork}(\;\cdot\;;\;\mA^{(n)})$ over all $n\times k$ label matrices. Let $\mC_n$ be the true one-hot label matrix associated with the DC-SBM class labels $\vY_n := [y_1, y_2, \ldots, y_n]$. Let $\mu$ be any $k\times k$ permutation matrix, and let $||\cdot||_F$ be the Frobenius norm. As in \cite{zhao2012consistency}, we consider two notions of consistency:
\begin{align}
    & &&\pP_n \left[\min_{\mu}||\hat{\mC}^\ast_n\mu - \mC_n||^2_F = 0\right] &&\rightarrow 1 &\text{as } n \rightarrow \infty && \text{(strong consistency)}\\
    &\forall \epsilon > 0, &&\pP_n \left[\min_{\mu}\frac{1}{n}||\hat{\mC}^\ast_n\mu - \mC_n||^2_F < \epsilon \right] &&\rightarrow 1 &\text{as } n \rightarrow \infty && \text{(weak consistency)}
\end{align}

We now state our main theorem, which shows that \thiswork's novel collapse constraint does not prevent asymptotically-consistent prediction of the optimal class labels.

\begin{theorem}
Under the conditions of Theorem 3.1 from \cite{zhao2012consistency}, when the cluster sizes of the DC-SBM are equal, $\mathcal{L}_{\thiswork}$ is strongly consistent when $\lambda_n/\log(n)\rightarrow\infty$ and weakly consistent when $\lambda_n\rightarrow\infty$. 
\end{theorem}

\begin{proof}
This result relies on a general theorem from \cite{zhao2012consistency}, which is restated here for completeness. For convenience, dependence on $n$ will be ignored when it does not cause ambiguity. The \thiswork objective can be re-written as a function of inter-class edge counts $O(\mA, y)_{rs} = \sum_{i,j}\mA_{ij}\cdot1(y_i=r,y_j=s)$ and class proportions $\pi(y)_r=n^{-1}\sum_i1(y_i=r)$. Let $\bar{O}(\mA, y)_r = \sum_{s}O(\mA, y)_{rs}$ and $m(\mA) = \sum_{r}\bar{O}(\mA, y)_r$. Henceforth dependence on $y$ will sometimes be ignored for clarity. With these notations, we have
\begin{equation}
    \mathcal{L}_{\thiswork}(C; \mA) = -\frac{1}{2m(\mA)}\sum_{r}\left[O(\mA, y)_{rr} - \frac{\bar{O}(\mA, y)_r^2}{m(\mA)}\right] + \frac{\sqrt{k}}{n}||n\pi(y)||_F - 1
\end{equation}
Hence $\mathcal{L}_{\thiswork}(C; \mA)$ can be written as a function $F(O, \pi)$ of a $k\times k$ edge count matrix $O$  and a $k\times 1$ class proportion vector $\pi$. Theorem 4.1 from \cite{zhao2012consistency} shows that if $F$ satisfies regularity conditions and is uniquely minimized by the conditional expectation of its arguments under the DC-SBM $\mP_n$, then the objective function that $F$ describes is both strongly and weakly consistent under the model.

The regularity conditions are listed in Section 4 of \cite{zhao2012consistency} and are easy to verify against $\mathcal{L}_{\thiswork}$. We now show that $\mathcal{L}_{\thiswork}$ is uniquely minimized by the conditional expectation of its arguments under the DC-SBM $\mP_n$. In other words, $\mathcal{L}_{\thiswork}$ must be uniquely minimized at any point $(y^\ast, \mA^\ast)$ such that $\mathbb{E}_n\left[\pi(y_n)\right] = \pi(y^\ast)$ and $\mathbb{E}_n\left[\mA^{(n)}\right] = \mA^\ast$. Note that the first term of $\mathcal{L}_{\thiswork}$ is the (negative) standard Newman-Girvan modularity, which was shown to be minimized under the expected DC-SBM in \cite{zhao2012consistency} (see the proof of Theorem 3.1). The proof is complete as the \thiswork collapse constraint is minimized when the cluster sizes are equal.
\end{proof}
 \section{Experiments}\label{sec:experiments}

In this section, we describe empirical experiments---involving both synthetic and real-world data---on \thiswork{} and baseline methods, to test robustness and performance against both graph clustering and label alignment metrics. We use open-source graph simulation tools, publicly-available datasets, and we release the implementation of \thiswork{} at this URL\footnote{\href{https://github.com/google-research/google-research/tree/master/graph_embedding/dmon}{\texttt{github.com/google-research/google-research/tree/master/graph\_embedding/dmon}}}.

\begin{wrapfigure}[12]{r}{7cm}
\vspace{-5mm} \footnotesize
\centering{
\newcolumntype{C}{>{\raggedleft\arraybackslash}X}
\newcolumntype{S}{>{\centering\arraybackslash\hsize=.5\hsize}X}
\caption{\label{tbl:datasets}Dataset statistics.}
\vspace*{-2mm}
\begin{tabularx}{\linewidth}{@{}p{1.85cm}CCSS@{}}
\toprule
\emph{dataset} & $|V|$ & $|E|$ & $|X|$ & $|Y|$ \\
\midrule
\dcora & 2708 & 5278 & 1433 & 7 \\
\dciteseer & 3327 & 4614 & 3703 & 6 \\
\dpubmed & 19717 & 44325 & 500 & 3 \\
\mbox{\damazoncomputers} & 13752 & 143604 & 767 & 10 \\
\mbox{\damazonphoto} & 7650 & 71831 & 745 & 8 \\
\mbox{\dcoauthoreng} & 14927 & 49305 & 4839 & 16 \\
\mbox{\dcoauthorcs} & 18333 & 81894 & 6805 & 15 \\
\mbox{\dcoauthorphy} & 34493 & 247962 & 8415 & 5 \\
\mbox{\dcoauthorchem} & 35409 & 157358 & 4877 & 14 \\
\mbox{\dcoauthormed} & 63282 & 810314 & 5538 & 17 \\
\mbox{\darxiv} & 169343 & 583121 & 128 & 40 \\
\bottomrule
\end{tabularx}}
\end{wrapfigure} 
\spara{Datasets.}
We use 11 real-world datasets for assessing model quality.
Cora, Citeseer, and Pubmed~\citep{sen2008} are citation networks; nodes represent papers connected by citation edges; features are bag-of-word abstracts, and labels represent paper topics.
\damazoncomputers and \damazonphoto~\citep{shchur2018} are subsets of the Amazon co-purchase graph for the computers and photo sections of the website, where nodes represent goods with edges between ones frequently purchased together; node features are bag-of-word reviews, and class labels are product category.
\antnew{\dcoauthorcs, \dcoauthorphy, \dcoauthormed, \dcoauthorchem, and \dcoauthoreng~\citep{shchur2018, shchur2019overlapping} are co-authorship networks based on the Microsoft Academic Graph for the computer science, physics, medicine, and engineering fields respectively; nodes are authors, which are connected by edge if they co-authored a paper together; node features are a collection of paper keywords for author’s papers; class
labels indicate most common fields of study.
\darxiv~\citep{hu2020open} is a paper co-citation dataset based on arXiv papers indexed by the Microsoft Academic graph. Nodes are papers; edges are citations, and class labels indicate the main category of the paper.}

\spara{Baselines.}
As we study how to leverage the information from both the graph and attributes, we employ two baselines that employ either strictly graph or attribute information.
We give a brief description of the baselines used below:

\begin{itemize}[leftmargin=0.2cm,itemindent=.3cm,labelwidth=\itemindent,labelsep=0cm,align=left,topsep=0pt,itemsep=-1ex,partopsep=0ex,parsep=0.75ex]
    \item \textbf{k-means(features)} is our baseline that only considers the feature data. We use the local Lloyd algorithm~\citep{lloyd1982} with the k-means++ seeding strategy~\citep{arthur2007}.
    \item \textbf{SBM}~\citep{peixoto2014} is a baseline that only relies on the graph structure. We estimate a constrained stochastic block model with given $k$, the \ant{maximum} number of clusters.
    \item \textbf{k-means(DeepWalk, features)}~\citep{perozzi2014} represents a na\"ive strategy of concatenating node attributes to learned node embeddings \ant{of the graph without attributes.}
    \item \textbf{k-means(DGI)}~\citep{velickovic2018} demonstrates the need of joint learning of clusters and representations. We learn unsupervised node representations \ant{of the attributed graph} with DGI and run k-means on the resulting representations.
    \antnew{\item \textbf{AGC(graph, features)}~\citep{zhang2019attributed} applied spectral clustering over features smoothed by the graph.
    \item \textbf{DAEGC(graph, features)}~\citep{wang2019attributed} is a graph reconstruction-based clustering method.
    \item \textbf{SDCN(graph, features)}~\citep{bo2020structural} is another graph reconstruction-based clustering method. Representations are initialized through k-means over auto-encoded features.
    \item \textbf{NOCD(graph, features)}~\citep{shchur2019overlapping} directly optimizes negative log-likelihood of the graph reconstruction of the Bernoulli-Poisson model.}
    \item \textbf{DiffPool(graph, features)}~\citep{ying2018} is an early graph pooling method.
    \item \textbf{MinCutPool(graph, features)}~\citep{bianchi2020} is a deep pooling method \ant{that orthogonalizes the cluster representations (cf.\ Section~\ref{eq:dmon-architecture} for discussion).}
    \item \textbf{Ortho(graph, features)}~\citep{bianchi2020} is a variant of \textbf{MinCutPool} \ant{that only does the cluster orthogonalization without \emph{any} graph-related objective}.
\end{itemize}

\spara{Metrics.}
We measure both the graph-based metrics of clustering and label correlation to study clustering performance of attributed graphs both in terms of graph and attribute structure.
\pal{For experiments on real-world data, we measure both (1) standard graph-based clustering metrics, and (2) correlation to ground-truth node labels. Our graph-based metrics are average cluster conductance (as per definition from~\citet{yang2015}) and graph modularity~\citep{newman2006}. Our label-based metrics are normalized mutual information (NMI) between the cluster assignments and labels and pairwise F1 score between all node pairs and their associated cluster pairs. For experiments on synthetic data, we report only the NMI against the (simulated) cluster labels.}
Where possible, we normalize all metrics by multiplying them by 100 for \ant{readability and} ease of comparison.

\spara{Parameter settings.}
We run \pal{both synthetic and real-world experiments} for 10 times and average results across runs.
All models were implemented in Tensorflow 2 and trained on CPUs.
We fix the architecture for all \pal{GNNs (including \thiswork{}) have one hidden layer---with 512 neurons for real-world data experiments, and 64 neurons for experiments with smaller synthetic graphs.}
We set the \ant{maximum} number of clusters to 16 for all datasets and methods.

\antnew{We aim to keep the hyperparameters constant across datasets, since (i) tuning hyperparameters in an unbiased way is non-trivial in the unsupervised setting, (ii) we report an array of metrics that relate to the graph and to the labels -- favoring one group over other is not in the scope of this paper.
One notable exceptions is SDCN, as it requires an autoencoder on the features to function. We keep the original three-layer encoder structure and set the architecture to $[32, 32, 128, 128, 32, 32]$ for synthetic SBM graphs and $[500, 500, 2000, 2000, 500, 500]$ for all other datasets. Setting hyperparameters in a more fair way to keep the total number of parameters same across different methods led to random performance.
}

\antnew{We now discuss method-specific hyperparameters for methods that have specific settings:
\begin{itemize}[leftmargin=0.2cm,itemindent=.3cm,labelwidth=\itemindent,labelsep=0cm,align=left,topsep=0pt,itemsep=-1ex,partopsep=0ex,parsep=0.75ex]
    \item \textbf{k-means(DeepWalk, features)}: We keep the learning parameters as per \citet{perozzi2014} number of walks $\gamma=80$, walk length $t=80$, and window size $w=10$.
    \item \textbf{AGC(graph, features)}: We set $k=1$ to mimic a single-layer graph convolution that we set for all GNN-based methods.
    \item \textbf{DAEGC(graph, features)}: We set the clustering loss coefficient $\gamma=10$ as per the original paper.
    \item \textbf{SDCN(graph, features)}: We set the clustering loss coefficient $\alpha=0.1$ and GNN loss coefficient $\beta=0.01$ as per the original paper.
    \item \textbf{NOCD(graph, features)}: We set the dropout to 0.5 uniformly across datasets and set the batch size to 2000 as per the original paper.
    \item \textbf{\thiswork(graph, features)}: We set the dropout to 0.5 uniformly across datasets.
\end{itemize}}

\antnew{The dropout setting we use does not benefit DAEC, SDCN, DiffPool, or MinCutPool, hence we are not able to unify the setting of that parameter across methods.
}

\subsection{Simulation Experiments on Stochastic Block Model with Features}\label{ssec:sbm-experiments}
\ant{To explore the robustness \pal{and sensitivity} of \pal{\thiswork{} and baselines} to variance in the graph and node features,} we \pal{conduct} a study on synthetic graphs using an \emph{attributed}, \emph{degree-corrected} stochastic block model (ADC-SBM).
The SBM~\citep{snijders1997estimation} plants a partition of clusters (``blocks'') in a graph, and generates edges via a distribution conditional on that partition.
This model has been used extensively to benchmark graph clustering methods~\citep{fortunato2016community}, and has recently been used for experiments on state-of-the-art \pal{supervised} GNNs~\citep{dwivedi2020benchmarking}.
In our version of the model, node features are also generated, using a multivariate mixture model, with the mixture memberships having some association (or de-assocation) with the cluster memberships.
\ant{We proceed to describing the graph generation and feature generation components of our model.}

\spara{ADC-SBM graph generation.} We fix a number of nodes $n$ and a number of clusters $k$, and choose node cluster memberships uniformly-at-random. Define the matrix $\mD_{k\times k}$ where $\mD_{ij}$ is the expected number of edges shared between nodes in clusters $i$ and $j$. We determine $\mD$ by fixing (1) the expected \emph{average} degree of the nodes $d\in \{1, n\}$, and (2) the expected \emph{average} sub-degree $d_{out} \leq d$ of a node to any cluster other than its own.
Note that the difference $d_{in} - d_{out}$, where $d_{in}:=d - d_{out}$, controls the spectral detectability of the clusters~\citep{nadakuditi2012graph}.
Finally, we generate a power-law $n$-vector $\theta$, where $\theta_i$ is proportional to $i$'s expected degree. We use the generated memberships and the generated parameters $\mD$ and $\theta$ as inputs to the degree-corrected SBM function from the \texttt{graph-tool}~\citep{peixoto_graph-tool_2014} package.

\begin{figure}[!t]
\begin{subfigure}[b]{0.24\linewidth}
\centering
\includegraphics[width=\linewidth]{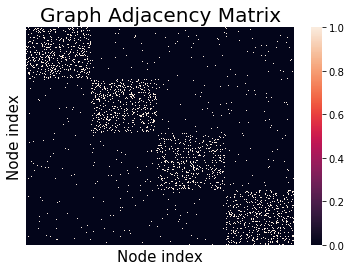}\caption{}
\end{subfigure}\hfill
\begin{subfigure}[b]{0.24\linewidth}
\centering
\includegraphics[width=\linewidth]{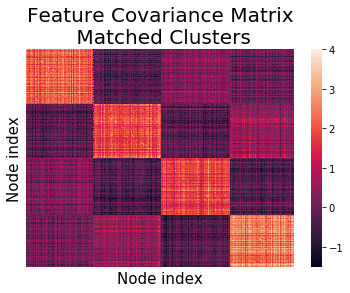}\caption{}
\end{subfigure}\hfill
\begin{subfigure}[b]{.24\linewidth}
\centering
\includegraphics[width=\linewidth]{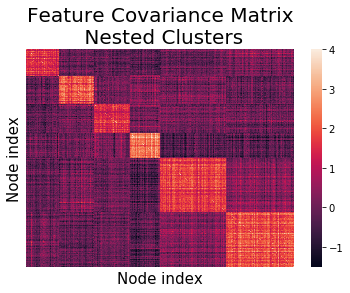}\caption{}
\end{subfigure}\hfill
\begin{subfigure}[b]{0.24\linewidth}
\centering
\includegraphics[width=\linewidth]{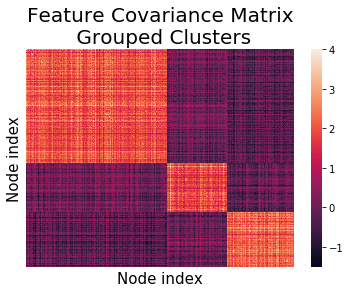}
\caption{}
\end{subfigure}
\caption{\label{fig:synthetic-illustration}\textbf{Illustration of synthetic data.} (a) 4-cluster graph adjacency matrix. (b) Covariance matrix of ``matched'' features: features that are clustered according to the graph clusters. (c) Covariance matrix of ``nested'' features: features that are clustered by incomplete nesting of the graph clusters. (d) Covariance matrix of ``grouped'' features: features that are clustered by incomplete grouping of the graph clusters.}
\end{figure} 
\begin{table}[!b]
\small
\centering
\newcolumntype{S}{>{\hsize=.075\hsize}X}
\newcolumntype{L}{>{\hsize=.24\hsize\raggedleft}X}
\caption{\label{tab:synthetic-scenarios} Synthetic ADC-SBM benchmark scenarios.}
\begin{tabularx}{\textwidth}{SLX}
\toprule
Scenario & Parameter & Description \\
\midrule
1 & $d_{out}\in[2.0, 5.0]$ & Increase graph cluster mixing signal. Higher = weaker clusters. \\
2 & $\sigma_c\in[10^{-2}, 10^1]$ & Increase feature cluster center variance. Higher = stronger clusters.\\
3 & $\sigma_c\in[10^{-2}, 10^1]$ & Increase feature cluster center variance, with nested feature clusters.\\
4 & $\sigma_c\in[10^{-2}, 10^1]$ & Increase feature cluster center variance, with grouped feature clusters.\\
5 & $d\in [2^2, 2^7]$ & Increase average degree. Higher = clearer graph signal.\\
6 & $d_{max} \in [2^2, 2^{10}]$ & Increase power law upper-bound. Higher = more extreme power law.\\
\bottomrule
    \end{tabularx}
\end{table} 
\spara{ADC-SBM feature generation.} We generate feature memberships from $k_f$ cluster labels. For graph clustering GNNs that operate both on edges and node features, it is important to examine performance on data where feature clusters diverge from or segment the graph clusters: thus potentially $k_f\ne k$. We examine cases where feature memberships \emph{match}, \emph{group}, or \emph{nest} the graph memberships, as illustrated in Figure~\ref{fig:synthetic-illustration}.
With feature memberships in-hand, we generate $k$ zero-mean feature cluster centers from a $s$-multivariate normal with covariance matrix $\sigma_c^2\cdot \mI_{s\times s}$.
Then, for feature cluster $i\leq k_f$, we generate its features from a $s$-multivariate normal with covariance matrix $\sigma^2\cdot \mI_{s\times s}$.
Note that the ratio $\sigma^2_c/\sigma^2$ controls the expected value of the classical between/within-sum-of-squares of the clusters.

The above paragraphs describe a single generation of our synthetic benchmark model, the ADC-SBM.
To study model robustness, we define ``default'' ADC-SBM parameters, and explore model parameters in a range around the defaults.
We configure our default model as follows: we generate graphs with $n=1,000$ nodes grouped in $k=4$ clusters, and $s=32$-dimensional features grouped in $k_f=4$ matching feature clusters with $\sigma=1$ intra-cluster center variance and $\sigma_c=3$ cluster center variance. We mimic real-world graphs' degree distribution with $d=20$ average degree and $d_{out}=2$ average inter-cluster degree with power law parameters $d_{min}=2, d_{max}=4, \alpha=2$.
In total, we consider 6 different scenarios, as described in Table~\ref{tab:synthetic-scenarios}.

\begin{figure}[htb]
\centering
\begin{tikzpicture}
\begin{groupplot}[group style={
                      group name=myplot,
                      group size= 3 by 2, horizontal sep=0.75cm,vertical sep=1.3cm},height=3.75cm,width=0.375 \linewidth,ymin=0,ymax=100,title style={at={(0.5,0.9)},anchor=south},every axis x label/.style={at={(axis description cs:0.5,-0.15)},anchor=north}]
\nextgroupplot[
 	title = \textbf{Scenario 1},
 	legend columns=6,
	legend style={at={(1.675,1.25)},anchor=south},
legend entries={\textbf{\thiswork{}}, DiffPool, MinCut, NOCD, SDCN, DAEGC},
	ylabel=NMI$\times100$,
	xlabel=$d_{out}$,
	xmin=1,
	xmax=5
]
\addplot[very thick,color=cycle2] table[x=param,y=mgcn2] {data/scenario1.tex};
\addplot[very thick,color=cycle7] table[x=param,y=diffpool] {data/scenario1.tex};
\addplot[very thick,color=cycle5] table[x=param,y=mincut] {data/scenario1.tex};
\addplot[very thick,color=cycle9] table[x=param,y=shchur] {data/scenario1.tex};
\addplot[very thick,color=cycle1] table[x=param,y=sdcn] {data/scenario1.tex};
\addplot[very thick,color=cycle11] table[x=param,y=daegc] {data/scenario1.tex};

\addplot[name path=mgcn_top_,color=cycle2!70,forget plot] table[x=param,y=mgcn2_high] {data/scenario1.tex};
\addplot[name path=mgcn_btm_,color=cycle2!70,forget plot] table[x=param,y=mgcn2_low] {data/scenario1.tex};
\addplot[cycle2!50,fill opacity=0.5,forget plot] fill between[of=mgcn_top_ and mgcn_btm_];

\addplot[name path=diffpool_top_,color=cycle7!70,forget plot] table[x=param,y=diffpool_high] {data/scenario1.tex};
\addplot[name path=diffpool_btm_,color=cycle7!70,forget plot] table[x=param,y=diffpool_low] {data/scenario1.tex};
\addplot[cycle7!50,fill opacity=0.5,forget plot] fill between[of=diffpool_top_ and diffpool_btm_];

\addplot[name path=mincut_top_,color=cycle5!70,forget plot] table[x=param,y=mincut_high] {data/scenario1.tex};
\addplot[name path=mincut_btm_,color=cycle5!70,forget plot] table[x=param,y=mincut_low] {data/scenario1.tex};
\addplot[cycle5!50,fill opacity=0.5,forget plot] fill between[of=mincut_top_ and mincut_btm_];

\addplot[name path=shchur_top_,color=cycle9!70,forget plot] table[x=param,y=shchur_high] {data/scenario1.tex};
\addplot[name path=shchur_btm_,color=cycle9!70,forget plot] table[x=param,y=shchur_low] {data/scenario1.tex};
\addplot[cycle9!50,fill opacity=0.5,forget plot] fill between[of=shchur_top_ and shchur_btm_];

\addplot[name path=sdcn_top_,color=cycle1!70,forget plot] table[x=param,y=sdcn_high] {data/scenario1.tex};
\addplot[name path=sdcn_btm_,color=cycle1!70,forget plot] table[x=param,y=sdcn_low] {data/scenario1.tex};
\addplot[cycle1!50,fill opacity=0.5,forget plot] fill between[of=sdcn_top_ and sdcn_btm_];

\addplot[name path=daegc_top_,color=cycle11!70,forget plot] table[x=param,y=daegc_high] {data/scenario1.tex};
\addplot[name path=daegc_btm_,color=cycle11!70,forget plot] table[x=param,y=daegc_low] {data/scenario1.tex};
\addplot[cycle11!50,fill opacity=0.5,forget plot] fill between[of=daegc_top_ and daegc_btm_];

\draw [color=cyclegray, ultra thick, draw opacity=0.75] (2.76393202250021,0) -- (2.76393202250021,100);
\draw[->,thick,color=black](axis cs:3,28.5)--(axis cs:2.85,40);
\node[anchor=east] (text) at (axis cs:3.9,20){\tiny \textbf{detectability limit}};
\nextgroupplot[
 	title = \textbf{Scenario 2},
	yticklabels={,,},
	xlabel=$\sigma_c$,
	xmode=log,
	xmin=0.01,
	xmax=10
]
\addplot[very thick,color=cycle2] table[x=param,y=mgcn2] {data/scenario2.tex};
\addplot[very thick,color=cycle7] table[x=param,y=diffpool] {data/scenario2.tex};
\addplot[very thick,color=cycle5] table[x=param,y=mincut] {data/scenario2.tex};
\addplot[very thick,color=cycle9] table[x=param,y=shchur] {data/scenario2.tex};
\addplot[very thick,color=cycle1] table[x=param,y=sdcn] {data/scenario2.tex};
\addplot[very thick,color=cycle11] table[x=param,y=daegc] {data/scenario2.tex};

\addplot[name path=mgcn_top_,color=cycle2!70,forget plot] table[x=param,y=mgcn2_high] {data/scenario2.tex};
\addplot[name path=mgcn_btm_,color=cycle2!70,forget plot] table[x=param,y=mgcn2_low] {data/scenario2.tex};
\addplot[cycle2!50,fill opacity=0.5,forget plot] fill between[of=mgcn_top_ and mgcn_btm_];

\addplot[name path=diffpool_top_,color=cycle7!70,forget plot] table[x=param,y=diffpool_high] {data/scenario2.tex};
\addplot[name path=diffpool_btm_,color=cycle7!70,forget plot] table[x=param,y=diffpool_low] {data/scenario2.tex};
\addplot[cycle7!50,fill opacity=0.5,forget plot] fill between[of=diffpool_top_ and diffpool_btm_];

\addplot[name path=mincut_top_,color=cycle5!70,forget plot] table[x=param,y=mincut_high] {data/scenario2.tex};
\addplot[name path=mincut_btm_,color=cycle5!70,forget plot] table[x=param,y=mincut_low] {data/scenario2.tex};
\addplot[cycle5!50,fill opacity=0.5,forget plot] fill between[of=mincut_top_ and mincut_btm_];

\addplot[name path=shchur_top_,color=cycle9!70,forget plot] table[x=param,y=shchur_high] {data/scenario2.tex};
\addplot[name path=shchur_btm_,color=cycle9!70,forget plot] table[x=param,y=shchur_low] {data/scenario2.tex};
\addplot[cycle9!50,fill opacity=0.5,forget plot] fill between[of=shchur_top_ and shchur_btm_];

\addplot[name path=sdcn_top_,color=cycle1!70,forget plot] table[x=param,y=sdcn_high] {data/scenario2.tex};
\addplot[name path=sdcn_btm_,color=cycle1!70,forget plot] table[x=param,y=sdcn_low] {data/scenario2.tex};
\addplot[cycle1!50,fill opacity=0.5,forget plot] fill between[of=sdcn_top_ and sdcn_btm_];

\addplot[name path=daegc_top_,color=cycle11!70,forget plot] table[x=param,y=daegc_high] {data/scenario2.tex};
\addplot[name path=daegc_btm_,color=cycle11!70,forget plot] table[x=param,y=daegc_low] {data/scenario2.tex};
\addplot[cycle11!50,fill opacity=0.5,forget plot] fill between[of=daegc_top_ and daegc_btm_];

\nextgroupplot[
 	title = \textbf{Scenario 3},
	yticklabels={,,},
	xlabel=$\sigma_c$,
	xmode=log,
	xmin=0.01,
	xmax=10
]
\addplot[very thick,color=cycle2] table[x=param,y=mgcn2] {data/scenario3.tex};
\addplot[very thick,color=cycle7] table[x=param,y=diffpool] {data/scenario3.tex};
\addplot[very thick,color=cycle5] table[x=param,y=mincut] {data/scenario3.tex};
\addplot[very thick,color=cycle9] table[x=param,y=shchur] {data/scenario3.tex};
\addplot[very thick,color=cycle1] table[x=param,y=sdcn] {data/scenario3.tex};
\addplot[very thick,color=cycle11] table[x=param,y=daegc] {data/scenario3.tex};

\addplot[name path=mgcn_top_,color=cycle2!70,forget plot] table[x=param,y=mgcn2_high] {data/scenario3.tex};
\addplot[name path=mgcn_btm_,color=cycle2!70,forget plot] table[x=param,y=mgcn2_low] {data/scenario3.tex};
\addplot[cycle2!50,fill opacity=0.5,forget plot] fill between[of=mgcn_top_ and mgcn_btm_];

\addplot[name path=diffpool_top_,color=cycle7!70,forget plot] table[x=param,y=diffpool_high] {data/scenario3.tex};
\addplot[name path=diffpool_btm_,color=cycle7!70,forget plot] table[x=param,y=diffpool_low] {data/scenario3.tex};
\addplot[cycle7!50,fill opacity=0.5,forget plot] fill between[of=diffpool_top_ and diffpool_btm_];

\addplot[name path=mincut_top_,color=cycle5!70,forget plot] table[x=param,y=mincut_high] {data/scenario3.tex};
\addplot[name path=mincut_btm_,color=cycle5!70,forget plot] table[x=param,y=mincut_low] {data/scenario3.tex};
\addplot[cycle5!50,fill opacity=0.5,forget plot] fill between[of=mincut_top_ and mincut_btm_];

\addplot[name path=shchur_top_,color=cycle9!70,forget plot] table[x=param,y=shchur_high] {data/scenario3.tex};
\addplot[name path=shchur_btm_,color=cycle9!70,forget plot] table[x=param,y=shchur_low] {data/scenario3.tex};
\addplot[cycle9!50,fill opacity=0.5,forget plot] fill between[of=shchur_top_ and shchur_btm_];

\addplot[name path=sdcn_top_,color=cycle1!70,forget plot] table[x=param,y=sdcn_high] {data/scenario3.tex};
\addplot[name path=sdcn_btm_,color=cycle1!70,forget plot] table[x=param,y=sdcn_low] {data/scenario3.tex};
\addplot[cycle1!50,fill opacity=0.5,forget plot] fill between[of=sdcn_top_ and sdcn_btm_];

\addplot[name path=daegc_top_,color=cycle11!70,forget plot] table[x=param,y=daegc_high] {data/scenario3.tex};
\addplot[name path=daegc_btm_,color=cycle11!70,forget plot] table[x=param,y=daegc_low] {data/scenario3.tex};
\addplot[cycle11!50,fill opacity=0.5,forget plot] fill between[of=daegc_top_ and daegc_btm_];

\nextgroupplot[
 	title = \textbf{Scenario 4},
	xlabel=$\sigma_c$,
	ylabel=NMI$\times100$,
	xmode=log,
	xmin=0.01,
	xmax=10
]
\addplot[very thick,color=cycle2] table[x=param,y=mgcn2] {data/scenario4.tex};
\addplot[very thick,color=cycle7] table[x=param,y=diffpool] {data/scenario4.tex};
\addplot[very thick,color=cycle5] table[x=param,y=mincut] {data/scenario4.tex};
\addplot[very thick,color=cycle9] table[x=param,y=shchur] {data/scenario4.tex};
\addplot[very thick,color=cycle1] table[x=param,y=sdcn] {data/scenario4.tex};
\addplot[very thick,color=cycle11] table[x=param,y=daegc] {data/scenario4.tex};

\addplot[name path=mgcn_top_,color=cycle2!70,forget plot] table[x=param,y=mgcn2_high] {data/scenario4.tex};
\addplot[name path=mgcn_btm_,color=cycle2!70,forget plot] table[x=param,y=mgcn2_low] {data/scenario4.tex};
\addplot[cycle2!50,fill opacity=0.5,forget plot] fill between[of=mgcn_top_ and mgcn_btm_];

\addplot[name path=diffpool_top_,color=cycle7!70,forget plot] table[x=param,y=diffpool_high] {data/scenario4.tex};
\addplot[name path=diffpool_btm_,color=cycle7!70,forget plot] table[x=param,y=diffpool_low] {data/scenario4.tex};
\addplot[cycle7!50,fill opacity=0.5,forget plot] fill between[of=diffpool_top_ and diffpool_btm_];

\addplot[name path=mincut_top_,color=cycle5!70,forget plot] table[x=param,y=mincut_high] {data/scenario4.tex};
\addplot[name path=mincut_btm_,color=cycle5!70,forget plot] table[x=param,y=mincut_low] {data/scenario4.tex};
\addplot[cycle5!50,fill opacity=0.5,forget plot] fill between[of=mincut_top_ and mincut_btm_];

\addplot[name path=shchur_top_,color=cycle9!70,forget plot] table[x=param,y=shchur_high] {data/scenario4.tex};
\addplot[name path=shchur_btm_,color=cycle9!70,forget plot] table[x=param,y=shchur_low] {data/scenario4.tex};
\addplot[cycle9!50,fill opacity=0.5,forget plot] fill between[of=shchur_top_ and shchur_btm_];

\addplot[name path=sdcn_top_,color=cycle1!70,forget plot] table[x=param,y=sdcn_high] {data/scenario4.tex};
\addplot[name path=sdcn_btm_,color=cycle1!70,forget plot] table[x=param,y=sdcn_low] {data/scenario4.tex};
\addplot[cycle4!50,fill opacity=0.5,forget plot] fill between[of=sdcn_top_ and sdcn_btm_];

\addplot[name path=daegc_top_,color=cycle11!70,forget plot] table[x=param,y=daegc_high] {data/scenario4.tex};
\addplot[name path=daegc_btm_,color=cycle11!70,forget plot] table[x=param,y=daegc_low] {data/scenario4.tex};
\addplot[cycle11!50,fill opacity=0.5,forget plot] fill between[of=daegc_top_ and daegc_btm_];

\nextgroupplot[
 	title = \textbf{Scenario 5},
	yticklabels={,,},
	xlabel=$d$,
	xmode=log,
	xmin=4,
	xmax=128,
	log basis x={2}
]
\addplot[very thick,color=cycle2] table[x=param,y=mgcn2] {data/scenario5.tex};
\addplot[very thick,color=cycle7] table[x=param,y=diffpool] {data/scenario5.tex};
\addplot[very thick,color=cycle5] table[x=param,y=mincut] {data/scenario5.tex};
\addplot[very thick,color=cycle9] table[x=param,y=shchur] {data/scenario5.tex};
\addplot[very thick,color=cycle1] table[x=param,y=sdcn] {data/scenario5.tex};
\addplot[very thick,color=cycle11] table[x=param,y=daegc] {data/scenario5.tex};

\addplot[name path=mgcn_top_,color=cycle2!70,forget plot] table[x=param,y=mgcn2_high] {data/scenario5.tex};
\addplot[name path=mgcn_btm_,color=cycle2!70,forget plot] table[x=param,y=mgcn2_low] {data/scenario5.tex};
\addplot[cycle2!50,fill opacity=0.5,forget plot] fill between[of=mgcn_top_ and mgcn_btm_];

\addplot[name path=diffpool_top_,color=cycle7!70,forget plot] table[x=param,y=diffpool_high] {data/scenario5.tex};
\addplot[name path=diffpool_btm_,color=cycle7!70,forget plot] table[x=param,y=diffpool_low] {data/scenario5.tex};
\addplot[cycle7!50,fill opacity=0.5,forget plot] fill between[of=diffpool_top_ and diffpool_btm_];

\addplot[name path=mincut_top_,color=cycle5!70,forget plot] table[x=param,y=mincut_high] {data/scenario5.tex};
\addplot[name path=mincut_btm_,color=cycle5!70,forget plot] table[x=param,y=mincut_low] {data/scenario5.tex};
\addplot[cycle5!50,fill opacity=0.5,forget plot] fill between[of=mincut_top_ and mincut_btm_];

\addplot[name path=shchur_top_,color=cycle9!70,forget plot] table[x=param,y=shchur_high] {data/scenario5.tex};
\addplot[name path=shchur_btm_,color=cycle9!70,forget plot] table[x=param,y=shchur_low] {data/scenario5.tex};
\addplot[cycle9!50,fill opacity=0.5,forget plot] fill between[of=shchur_top_ and shchur_btm_];

\addplot[name path=sdcn_top_,color=cycle1!70,forget plot] table[x=param,y=sdcn_high] {data/scenario5.tex};
\addplot[name path=sdcn_btm_,color=cycle1!70,forget plot] table[x=param,y=sdcn_low] {data/scenario5.tex};
\addplot[cycle1!50,fill opacity=0.5,forget plot] fill between[of=sdcn_top_ and sdcn_btm_];

\addplot[name path=daegc_top_,color=cycle11!70,forget plot] table[x=param,y=daegc_high] {data/scenario5.tex};
\addplot[name path=daegc_btm_,color=cycle11!70,forget plot] table[x=param,y=daegc_low] {data/scenario5.tex};
\addplot[cycle11!50,fill opacity=0.5,forget plot] fill between[of=daegc_top_ and daegc_btm_];

\nextgroupplot[
 	title = \textbf{Scenario 6},
	yticklabels={,,},
	xlabel=$d_{\max}$,
	xmode=log,
	xmin=4,
	xmax=1024,
	log basis x={2},
    max space between ticks=20,
]
\addplot[very thick,color=cycle2] table[x=param,y=mgcn2] {data/scenario6.tex};
\addplot[very thick,color=cycle7] table[x=param,y=diffpool] {data/scenario6.tex};
\addplot[very thick,color=cycle5] table[x=param,y=mincut] {data/scenario6.tex};
\addplot[very thick,color=cycle9] table[x=param,y=shchur] {data/scenario6.tex};
\addplot[very thick,color=cycle1] table[x=param,y=sdcn] {data/scenario6.tex};
\addplot[very thick,color=cycle11] table[x=param,y=daegc] {data/scenario6.tex};

\addplot[name path=mgcn_top_,color=cycle2!70,forget plot] table[x=param,y=mgcn2_high] {data/scenario6.tex};
\addplot[name path=mgcn_btm_,color=cycle2!70,forget plot] table[x=param,y=mgcn2_low] {data/scenario6.tex};
\addplot[cycle2!50,fill opacity=0.5,forget plot] fill between[of=mgcn_top_ and mgcn_btm_];

\addplot[name path=diffpool_top_,color=cycle7!70,forget plot] table[x=param,y=diffpool_high] {data/scenario6.tex};
\addplot[name path=diffpool_btm_,color=cycle7!70,forget plot] table[x=param,y=diffpool_low] {data/scenario6.tex};
\addplot[cycle7!50,fill opacity=0.5,forget plot] fill between[of=diffpool_top_ and diffpool_btm_];

\addplot[name path=mincut_top_,color=cycle5!70,forget plot] table[x=param,y=mincut_high] {data/scenario6.tex};
\addplot[name path=mincut_btm_,color=cycle5!70,forget plot] table[x=param,y=mincut_low] {data/scenario6.tex};
\addplot[cycle5!50,fill opacity=0.5,forget plot] fill between[of=mincut_top_ and mincut_btm_];

\addplot[name path=shchur_top_,color=cycle9!70,forget plot] table[x=param,y=shchur_high] {data/scenario6.tex};
\addplot[name path=shchur_btm_,color=cycle9!70,forget plot] table[x=param,y=shchur_low] {data/scenario6.tex};
\addplot[cycle9!50,fill opacity=0.5,forget plot] fill between[of=shchur_top_ and shchur_btm_];

\addplot[name path=sdcn_top_,color=cycle1!70,forget plot] table[x=param,y=sdcn_high] {data/scenario6.tex};
\addplot[name path=sdcn_btm_,color=cycle1!70,forget plot] table[x=param,y=sdcn_low] {data/scenario6.tex};
\addplot[cycle1!50,fill opacity=0.5,forget plot] fill between[of=sdcn_top_ and sdcn_btm_];

\addplot[name path=daegc_top_,color=cycle11!70,forget plot] table[x=param,y=daegc_high] {data/scenario6.tex};
\addplot[name path=daegc_btm_,color=cycle11!70,forget plot] table[x=param,y=daegc_low] {data/scenario6.tex};
\addplot[cycle11!50,fill opacity=0.5,forget plot] fill between[of=daegc_top_ and daegc_btm_];
\end{groupplot}
\end{tikzpicture}
\caption{Synthetic results on the ADC-SBM model with 6 different scenarios described in Table~\ref{tab:synthetic-scenarios}. \ant{\thiswork{} significantly outperforms other neural graph pooling method baselines.}}\label{fig:synthetic-results-pooling}
\end{figure}

\begin{figure}[htb]
\centering
\begin{tikzpicture}
\begin{groupplot}[group style={
                      group name=myplot,
                      group size= 3 by 2, horizontal sep=0.75cm,vertical sep=1.3cm},height=3.75cm,width=0.375 \linewidth,ymin=0,ymax=100,title style={at={(0.5,0.9)},anchor=south},every axis x label/.style={at={(axis description cs:0.5,-0.15)},anchor=north}]
\nextgroupplot[
 	title = \textbf{Scenario 1},
 	legend columns=6,
	legend style={at={(1.675,1.25)},anchor=south},
legend entries={\textbf{\thiswork{}}, SBM, k-m(DGI), k-m(DW), k-m(feat.), AGC},
	ylabel=NMI$\times100$,
	xlabel=$d_{out}$,
	xmin=1,
	xmax=5
]
\addplot[very thick,color=cycle2] table[x=param,y=mgcn2] {data/scenario1.tex};
\addplot[very thick,color=cycle3] table[x=param,y=sbm] {data/scenario1.tex};
\addplot[very thick,color=cycle6] table[x=param,y=dgi] {data/scenario1.tex};
\addplot[very thick,color=cycle8] table[x=param,y=deepwalk] {data/scenario1.tex};
\addplot[very thick,color=cycle4] table[x=param,y=km] {data/scenario1.tex};
\addplot[very thick,color=cycle11] table[x=param,y=agc] {data/scenario1.tex};

\addplot[name path=mgcn_top_,color=cycle2!70,forget plot] table[x=param,y=mgcn2_high] {data/scenario1.tex};
\addplot[name path=mgcn_btm_,color=cycle2!70,forget plot] table[x=param,y=mgcn2_low] {data/scenario1.tex};
\addplot[cycle2!50,fill opacity=0.5,forget plot] fill between[of=mgcn_top_ and mgcn_btm_];

\addplot[name path=sbm_top_,color=cycle3!70,forget plot] table[x=param,y=sbm_high] {data/scenario1.tex};
\addplot[name path=sbm_btm_,color=cycle3!70,forget plot] table[x=param,y=sbm_low] {data/scenario1.tex};
\addplot[cycle3!50,fill opacity=0.5,forget plot] fill between[of=sbm_top_ and sbm_btm_];

\addplot[name path=km_top_,color=cycle4!70,forget plot] table[x=param,y=km_high] {data/scenario1.tex};
\addplot[name path=km_btm_,color=cycle4!70,forget plot] table[x=param,y=km_low] {data/scenario1.tex};
\addplot[cycle4!50,fill opacity=0.5,forget plot] fill between[of=km_top_ and km_btm_];

\addplot[name path=deepwalk_top_,color=cycle8!70,forget plot] table[x=param,y=deepwalk_high] {data/scenario1.tex};
\addplot[name path=deepwalk_btm_,color=cycle8!70,forget plot] table[x=param,y=deepwalk_low] {data/scenario1.tex};
\addplot[cycle8!50,fill opacity=0.5,forget plot] fill between[of=deepwalk_top_ and deepwalk_btm_];

\addplot[name path=dgi_top_,color=cycle6!70,forget plot] table[x=param,y=dgi_high] {data/scenario1.tex};
\addplot[name path=dgi_btm_,color=cycle6!70,forget plot] table[x=param,y=dgi_low] {data/scenario1.tex};
\addplot[cycle6!50,fill opacity=0.5,forget plot] fill between[of=dgi_top_ and dgi_btm_];

\addplot[name path=agc_top_,color=cycle11!70,forget plot] table[x=param,y=agc_high] {data/scenario1.tex};
\addplot[name path=agc_btm_,color=cycle11!70,forget plot] table[x=param,y=agc_low] {data/scenario1.tex};
\addplot[cycle10!50,fill opacity=0.5,forget plot] fill between[of=agc_top_ and agc_btm_];

\draw [color=cyclegray, ultra thick, draw opacity=0.75] (2.76393202250021,0) -- (2.76393202250021,100);
\draw[->,thick,color=black](axis cs:3,28.5)--(axis cs:2.85,40);
\node[anchor=east] (text) at (axis cs:3.9,20){\tiny \textbf{detectability limit}};
\nextgroupplot[
 	title = \textbf{Scenario 2},
	yticklabels={,,},
	xlabel=$\sigma_c$,
	xmode=log,
	xmin=0.01,
	xmax=10
]
\addplot[very thick,color=cycle2] table[x=param,y=mgcn2] {data/scenario2.tex};
\addplot[very thick,color=cycle3] table[x=param,y=sbm] {data/scenario2.tex};
\addplot[very thick,color=cycle4] table[x=param,y=km] {data/scenario2.tex};
\addplot[very thick,color=cycle8] table[x=param,y=deepwalk] {data/scenario2.tex};
\addplot[very thick,color=cycle6] table[x=param,y=dgi] {data/scenario2.tex};
\addplot[very thick,color=cycle11] table[x=param,y=agc] {data/scenario2.tex};

\addplot[name path=mgcn_top_,color=cycle2!70,forget plot] table[x=param,y=mgcn2_high] {data/scenario2.tex};
\addplot[name path=mgcn_btm_,color=cycle2!70,forget plot] table[x=param,y=mgcn2_low] {data/scenario2.tex};
\addplot[cycle2!50,fill opacity=0.5,forget plot] fill between[of=mgcn_top_ and mgcn_btm_];

\addplot[name path=sbm_top_,color=cycle3!70,forget plot] table[x=param,y=sbm_high] {data/scenario2.tex};
\addplot[name path=sbm_btm_,color=cycle3!70,forget plot] table[x=param,y=sbm_low] {data/scenario2.tex};
\addplot[cycle3!50,fill opacity=0.5,forget plot] fill between[of=sbm_top_ and sbm_btm_];

\addplot[name path=km_top_,color=cycle4!70,forget plot] table[x=param,y=km_high] {data/scenario2.tex};
\addplot[name path=km_btm_,color=cycle4!70,forget plot] table[x=param,y=km_low] {data/scenario2.tex};
\addplot[cycle4!50,fill opacity=0.5,forget plot] fill between[of=km_top_ and km_btm_];

\addplot[name path=deepwalk_top_,color=cycle8!70,forget plot] table[x=param,y=deepwalk_high] {data/scenario2.tex};
\addplot[name path=deepwalk_btm_,color=cycle8!70,forget plot] table[x=param,y=deepwalk_low] {data/scenario2.tex};
\addplot[cycle8!50,fill opacity=0.5,forget plot] fill between[of=deepwalk_top_ and deepwalk_btm_];

\addplot[name path=dgi_top_,color=cycle6!70,forget plot] table[x=param,y=dgi_high] {data/scenario2.tex};
\addplot[name path=dgi_btm_,color=cycle6!70,forget plot] table[x=param,y=dgi_low] {data/scenario2.tex};
\addplot[cycle6!50,fill opacity=0.5,forget plot] fill between[of=dgi_top_ and dgi_btm_];

\addplot[name path=agc_top_,color=cycle11!70,forget plot] table[x=param,y=agc_high] {data/scenario2.tex};
\addplot[name path=agc_btm_,color=cycle11!70,forget plot] table[x=param,y=agc_low] {data/scenario2.tex};
\addplot[cycle10!50,fill opacity=0.5,forget plot] fill between[of=agc_top_ and agc_btm_];
\nextgroupplot[
 	title = \textbf{Scenario 3},
	yticklabels={,,},
	xlabel=$\sigma_c$,
	xmode=log,
	xmin=0.01,
	xmax=10
]
\addplot[very thick,color=cycle2] table[x=param,y=mgcn2] {data/scenario3.tex};
\addplot[very thick,color=cycle3] table[x=param,y=sbm] {data/scenario3.tex};
\addplot[very thick,color=cycle4] table[x=param,y=km] {data/scenario3.tex};
\addplot[very thick,color=cycle8] table[x=param,y=deepwalk] {data/scenario3.tex};
\addplot[very thick,color=cycle6] table[x=param,y=dgi] {data/scenario3.tex};
\addplot[very thick,color=cycle11] table[x=param,y=agc] {data/scenario3.tex};

\addplot[name path=mgcn_top_,color=cycle2!70,forget plot] table[x=param,y=mgcn2_high] {data/scenario3.tex};
\addplot[name path=mgcn_btm_,color=cycle2!70,forget plot] table[x=param,y=mgcn2_low] {data/scenario3.tex};
\addplot[cycle2!50,fill opacity=0.5,forget plot] fill between[of=mgcn_top_ and mgcn_btm_];

\addplot[name path=sbm_top_,color=cycle3!70,forget plot] table[x=param,y=sbm_high] {data/scenario3.tex};
\addplot[name path=sbm_btm_,color=cycle3!70,forget plot] table[x=param,y=sbm_low] {data/scenario3.tex};
\addplot[cycle3!50,fill opacity=0.5,forget plot] fill between[of=sbm_top_ and sbm_btm_];

\addplot[name path=km_top_,color=cycle4!70,forget plot] table[x=param,y=km_high] {data/scenario3.tex};
\addplot[name path=km_btm_,color=cycle4!70,forget plot] table[x=param,y=km_low] {data/scenario3.tex};
\addplot[cycle4!50,fill opacity=0.5,forget plot] fill between[of=km_top_ and km_btm_];

\addplot[name path=deepwalk_top_,color=cycle8!70,forget plot] table[x=param,y=deepwalk_high] {data/scenario3.tex};
\addplot[name path=deepwalk_btm_,color=cycle8!70,forget plot] table[x=param,y=deepwalk_low] {data/scenario3.tex};
\addplot[cycle8!50,fill opacity=0.5,forget plot] fill between[of=deepwalk_top_ and deepwalk_btm_];

\addplot[name path=dgi_top_,color=cycle6!70,forget plot] table[x=param,y=dgi_high] {data/scenario3.tex};
\addplot[name path=dgi_btm_,color=cycle6!70,forget plot] table[x=param,y=dgi_low] {data/scenario3.tex};
\addplot[cycle6!50,fill opacity=0.5,forget plot] fill between[of=dgi_top_ and dgi_btm_];

\addplot[name path=agc_top_,color=cycle11!70,forget plot] table[x=param,y=agc_high] {data/scenario3.tex};
\addplot[name path=agc_btm_,color=cycle11!70,forget plot] table[x=param,y=agc_low] {data/scenario3.tex};
\addplot[cycle10!50,fill opacity=0.5,forget plot] fill between[of=agc_top_ and agc_btm_];
\nextgroupplot[
 	title = \textbf{Scenario 4},
	xlabel=$\sigma_c$,
	ylabel=NMI$\times100$,
	xmode=log,
	xmin=0.01,
	xmax=10
]
\addplot[very thick,color=cycle2] table[x=param,y=mgcn2] {data/scenario4.tex};
\addplot[very thick,color=cycle3] table[x=param,y=sbm] {data/scenario4.tex};
\addplot[very thick,color=cycle4] table[x=param,y=km] {data/scenario4.tex};
\addplot[very thick,color=cycle8] table[x=param,y=deepwalk] {data/scenario4.tex};
\addplot[very thick,color=cycle6] table[x=param,y=dgi] {data/scenario4.tex};
\addplot[very thick,color=cycle11] table[x=param,y=agc] {data/scenario4.tex};

\addplot[name path=mgcn_top_,color=cycle2!70,forget plot] table[x=param,y=mgcn2_high] {data/scenario4.tex};
\addplot[name path=mgcn_btm_,color=cycle2!70,forget plot] table[x=param,y=mgcn2_low] {data/scenario4.tex};
\addplot[cycle2!50,fill opacity=0.5,forget plot] fill between[of=mgcn_top_ and mgcn_btm_];

\addplot[name path=sbm_top_,color=cycle3!70,forget plot] table[x=param,y=sbm_high] {data/scenario4.tex};
\addplot[name path=sbm_btm_,color=cycle3!70,forget plot] table[x=param,y=sbm_low] {data/scenario4.tex};
\addplot[cycle3!50,fill opacity=0.5,forget plot] fill between[of=sbm_top_ and sbm_btm_];

\addplot[name path=km_top_,color=cycle4!70,forget plot] table[x=param,y=km_high] {data/scenario4.tex};
\addplot[name path=km_btm_,color=cycle4!70,forget plot] table[x=param,y=km_low] {data/scenario4.tex};
\addplot[cycle4!50,fill opacity=0.5,forget plot] fill between[of=km_top_ and km_btm_];

\addplot[name path=deepwalk_top_,color=cycle8!70,forget plot] table[x=param,y=deepwalk_high] {data/scenario4.tex};
\addplot[name path=deepwalk_btm_,color=cycle8!70,forget plot] table[x=param,y=deepwalk_low] {data/scenario4.tex};
\addplot[cycle8!50,fill opacity=0.5,forget plot] fill between[of=deepwalk_top_ and deepwalk_btm_];

\addplot[name path=dgi_top_,color=cycle6!70,forget plot] table[x=param,y=dgi_high] {data/scenario4.tex};
\addplot[name path=dgi_btm_,color=cycle6!70,forget plot] table[x=param,y=dgi_low] {data/scenario4.tex};
\addplot[cycle6!50,fill opacity=0.5,forget plot] fill between[of=dgi_top_ and dgi_btm_];

\addplot[name path=agc_top_,color=cycle11!70,forget plot] table[x=param,y=agc_high] {data/scenario4.tex};
\addplot[name path=agc_btm_,color=cycle11!70,forget plot] table[x=param,y=agc_low] {data/scenario4.tex};
\addplot[cycle10!50,fill opacity=0.5,forget plot] fill between[of=agc_top_ and agc_btm_];
\nextgroupplot[
 	title = \textbf{Scenario 5},
	yticklabels={,,},
	xlabel=$d$,
	xmode=log,
	xmin=4,
	xmax=128,
	log basis x={2}
]
\addplot[very thick,color=cycle2] table[x=param,y=mgcn2] {data/scenario5.tex};
\addplot[very thick,color=cycle3] table[x=param,y=sbm] {data/scenario5.tex};
\addplot[very thick,color=cycle4] table[x=param,y=km] {data/scenario5.tex};
\addplot[very thick,color=cycle8] table[x=param,y=deepwalk] {data/scenario5.tex};
\addplot[very thick,color=cycle6] table[x=param,y=dgi] {data/scenario5.tex};
\addplot[very thick,color=cycle11] table[x=param,y=agc] {data/scenario5.tex};

\addplot[name path=mgcn_top_,color=cycle2!70,forget plot] table[x=param,y=mgcn2_high] {data/scenario5.tex};
\addplot[name path=mgcn_btm_,color=cycle2!70,forget plot] table[x=param,y=mgcn2_low] {data/scenario5.tex};
\addplot[cycle2!50,fill opacity=0.5,forget plot] fill between[of=mgcn_top_ and mgcn_btm_];

\addplot[name path=sbm_top_,color=cycle3!70,forget plot] table[x=param,y=sbm_high] {data/scenario5.tex};
\addplot[name path=sbm_btm_,color=cycle3!70,forget plot] table[x=param,y=sbm_low] {data/scenario5.tex};
\addplot[cycle3!50,fill opacity=0.5,forget plot] fill between[of=sbm_top_ and sbm_btm_];

\addplot[name path=km_top_,color=cycle4!70,forget plot] table[x=param,y=km_high] {data/scenario5.tex};
\addplot[name path=km_btm_,color=cycle4!70,forget plot] table[x=param,y=km_low] {data/scenario5.tex};
\addplot[cycle4!50,fill opacity=0.5,forget plot] fill between[of=km_top_ and km_btm_];

\addplot[name path=deepwalk_top_,color=cycle8!70,forget plot] table[x=param,y=deepwalk_high] {data/scenario5.tex};
\addplot[name path=deepwalk_btm_,color=cycle8!70,forget plot] table[x=param,y=deepwalk_low] {data/scenario5.tex};
\addplot[cycle8!50,fill opacity=0.5,forget plot] fill between[of=deepwalk_top_ and deepwalk_btm_];

\addplot[name path=dgi_top_,color=cycle6!70,forget plot] table[x=param,y=dgi_high] {data/scenario5.tex};
\addplot[name path=dgi_btm_,color=cycle6!70,forget plot] table[x=param,y=dgi_low] {data/scenario5.tex};
\addplot[cycle6!50,fill opacity=0.5,forget plot] fill between[of=dgi_top_ and dgi_btm_];

\addplot[name path=agc_top_,color=cycle11!70,forget plot] table[x=param,y=agc_high] {data/scenario5.tex};
\addplot[name path=agc_btm_,color=cycle11!70,forget plot] table[x=param,y=agc_low] {data/scenario5.tex};
\addplot[cycle10!50,fill opacity=0.5,forget plot] fill between[of=agc_top_ and agc_btm_];
\nextgroupplot[
 	title = \textbf{Scenario 6},
	yticklabels={,,},
	xlabel=$d_{\max}$,
	xmode=log,
	xmin=4,
	xmax=1024,
	log basis x={2},
    max space between ticks=20,
]
\addplot[very thick,color=cycle2] table[x=param,y=mgcn2] {data/scenario6.tex};
\addplot[very thick,color=cycle3] table[x=param,y=sbm] {data/scenario6.tex};
\addplot[very thick,color=cycle4] table[x=param,y=km] {data/scenario6.tex};
\addplot[very thick,color=cycle8] table[x=param,y=deepwalk] {data/scenario6.tex};
\addplot[very thick,color=cycle6] table[x=param,y=dgi] {data/scenario6.tex};
\addplot[very thick,color=cycle11] table[x=param,y=agc] {data/scenario6.tex};

\addplot[name path=mgcn_top_,color=cycle2!70,forget plot] table[x=param,y=mgcn2_high] {data/scenario6.tex};
\addplot[name path=mgcn_btm_,color=cycle2!70,forget plot] table[x=param,y=mgcn2_low] {data/scenario6.tex};
\addplot[cycle2!50,fill opacity=0.5,forget plot] fill between[of=mgcn_top_ and mgcn_btm_];

\addplot[name path=sbm_top_,color=cycle3!70,forget plot] table[x=param,y=sbm_high] {data/scenario6.tex};
\addplot[name path=sbm_btm_,color=cycle3!70,forget plot] table[x=param,y=sbm_low] {data/scenario6.tex};
\addplot[cycle3!50,fill opacity=0.5,forget plot] fill between[of=sbm_top_ and sbm_btm_];

\addplot[name path=km_top_,color=cycle4!70,forget plot] table[x=param,y=km_high] {data/scenario6.tex};
\addplot[name path=km_btm_,color=cycle4!70,forget plot] table[x=param,y=km_low] {data/scenario6.tex};
\addplot[cycle4!50,fill opacity=0.5,forget plot] fill between[of=km_top_ and km_btm_];

\addplot[name path=deepwalk_top_,color=cycle8!70,forget plot] table[x=param,y=deepwalk_high] {data/scenario6.tex};
\addplot[name path=deepwalk_btm_,color=cycle8!70,forget plot] table[x=param,y=deepwalk_low] {data/scenario6.tex};
\addplot[cycle8!50,fill opacity=0.5,forget plot] fill between[of=deepwalk_top_ and deepwalk_btm_];

\addplot[name path=dgi_top_,color=cycle6!70,forget plot] table[x=param,y=dgi_high] {data/scenario6.tex};
\addplot[name path=dgi_btm_,color=cycle6!70,forget plot] table[x=param,y=dgi_low] {data/scenario6.tex};
\addplot[cycle6!50,fill opacity=0.5,forget plot] fill between[of=dgi_top_ and dgi_btm_];

\addplot[name path=agc_top_,color=cycle11!70,forget plot] table[x=param,y=agc_high] {data/scenario6.tex};
\addplot[name path=agc_btm_,color=cycle11!70,forget plot] table[x=param,y=agc_low] {data/scenario6.tex};
\addplot[cycle10!50,fill opacity=0.5,forget plot] fill between[of=agc_top_ and agc_btm_];
\end{groupplot}
\end{tikzpicture}
\caption{Synthetic results on the ADC-SBM model with 6 different scenarios described in Table~\ref{tab:synthetic-scenarios}. \ant{\thiswork{} leverages information from both graph structure and node attributes.}}\label{fig:synthetic-results-baselines}
\end{figure}
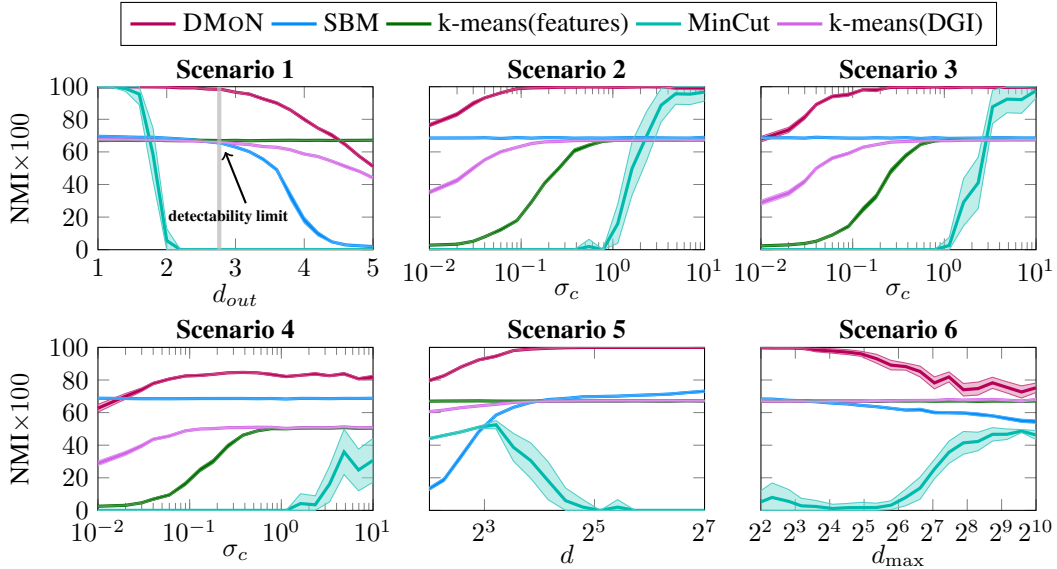

\mpara{Results.}
\ant{We split the presentation of results for neural methods (Figure~\ref{fig:synthetic-results-pooling}) and other baselines (Figure~\ref{fig:synthetic-results-baselines}) for the ease of understanding.
Overall, end-to-end graph pooling methods outperform multi-step baselines by a wide margin.
\thiswork demonstrates overwhelming superiority over all baselines, with the most significant improvements over other pooling methods. MinCut pooling suffers from the presence of even the weak noise in the graph (Scenario~1) or in the features (Scenario~2).
Moreover, it is susceptible to both nested and grouped features (Scenarios~3~and~4), while \thiswork and DiffPool are less sensitive to these variations.
Both DiffPool and MinCutPool are dependent on the sparsity level and degree homogeneity of the graph --- DiffPool performs better on denser graphs while MinCutPool shows the opposite picture.
}

\begin{table}[htb]
\footnotesize
\centering{
\newcolumntype{R}{>{\raggedleft\arraybackslash}X}
\newcolumntype{C}{>{\centering\arraybackslash}X}
\caption{Results on three datasets from~\cite{sen2008} in terms of graph conductance $\con$, modularity $\modu$, NMI with ground-truth labels, and pairwise F1 measure. \ant{We group the methods into three categories: baselines using only one aspect of data, neural representation learning, and neural graph pooling methods.} We highlight best neural method performance.}
\begin{tabularx}{\linewidth}{p{2cm}CCCCCCCCCCCC}
\toprule
\multicolumn{1}{C}{} & \multicolumn{4}{c}{\textbf{\dcora}} & \multicolumn{4}{c}{\textbf{\dciteseer}} & \multicolumn{4}{c}{\textbf{\dpubmed}} \\
\multicolumn{1}{C}{} & \multicolumn{2}{c}{\emph{graph}} & \multicolumn{2}{c}{\emph{labels}} & \multicolumn{2}{c}{\emph{graph}} & \multicolumn{2}{c}{\emph{labels}} & \multicolumn{2}{c}{\emph{graph}} & \multicolumn{2}{c}{\emph{labels}}  \\
\cmidrule(lr){2-5}\cmidrule(lr){6-9}\cmidrule(lr){10-13}
\emph{method} & $\con\downarrow$ & $\modu\uparrow$ & NMI$\uparrow$ & F1$\uparrow$ & $\con\downarrow$ & $\modu\uparrow$ & NMI$\uparrow$ & F1$\uparrow$ & $\con\downarrow$ & $\modu\uparrow$ & NMI$\uparrow$ & F1$\uparrow$\\
\midrule
\mbox{k-m(feat)} & 61.7 & 19.8 & 18.5 & 27.0 & 60.5 & 30.3 & 24.5 & 29.2 & 55.8 & 33.4 & 19.4 & 24.4 \\
\mbox{SBM} & 15.4 & 77.3 & 36.2 & 30.2 & 14.2 & 78.1 & 15.3 & 19.1 & 39.0 & 53.5 & 16.4 & 16.7 \\
\midrule
\mbox{k-m(DW)} & 62.1 & 30.7 & 24.3 & 24.8 & 68.1 & 24.3 & 27.6 & 24.8 & 16.6 & 75.3 & 22.9 & 17.2 \\
\mbox{AGC} & 48.9 & 43.2 & 34.1 & 28.9 & 41.9 & 50.0 & 25.5 & 27.5 & 44.9 & 46.8 & 18.2 & 18.4 \\
\mbox{SDCN} & 37.5 & 50.8 & 27.9 & 29.9 & 20.0 & 62.3 & 31.4 & 41.9 & 22.4 & 50.3 & 19.5 & 29.9 \\
\mbox{DAEGC} & 56.8 & 33.5 & 8.3 & 13.6 & 47.6 & 36.4 & 4.3 & 18.0 & 53.6 & 37.5 & 4.4 & 11.6 \\
\mbox{k-m(DGI)} & 28.0 & 64.0 & \textbf{52.7} & 40.1 & 17.5 & 73.7 & \textbf{40.4} & 39.4 & 82.9 & 9.6 & 22.0 & 26.4 \\
\midrule
\mbox{NOCD} & 14.7 & \textbf{78.3} & 46.3 & 36.7 & 6.8 &  \textbf{84.4} & 20.0 & 24.1 & 21.7 &  \textbf{69.6} & 25.5 & 20.8 \\
\mbox{DiffPool} & 26.1 & 66.3 & 32.9 & 34.4 & 26.0 & 63.4 & 20.0 & 23.5 & 32.9 & 56.8 & 20.2 & 26.3 \\
\mbox{MinCut} & 23.3 & 70.3 & 35.8 & 25.0 & 14.1 & 78.9 & 25.9 & 20.1 & 29.6 & 63.1 & 25.4 & 15.8 \\
\mbox{Ortho} & 28.0 & 65.6 & 38.4 & 26.6 & 18.4 & 74.5 & 26.1 & 20.5 & 57.8 & 32.9 & 20.3 & 13.9 \\
\midrule
\mbox{\thiswork} & \textbf{12.2} & 76.5 & 48.8 & \textbf{48.8} & \textbf{5.1} &79.3 & 33.7 & \textbf{43.2} & \textbf{17.7} & 65.4 & \textbf{29.8} & \textbf{33.9} \\
\bottomrule
\end{tabularx}}
\label{tbl:clustering}
\end{table} 
\ant{To better understand the limits of the task, we study the performance of our baselines and report the results on Figure~\ref{fig:synthetic-results-baselines}.
In particular our interest lies in the performance of the SBM and pure k-means over features, as these two baselines depict the performance possible when utilizing only one aspect of the data.
Scenario 1 shows that \thiswork{} can effectively leverage the feature signal to obtain outstanding clustering performance even when the graph structure is close to random, far beyond the spectral detectability threshold (pictured in gray).
Scenario 2 demonstrates that even in the presence of a weak feature signal \thiswork outperforms stochastic SBM minimization.
We also notice that while the k-means(DGI) baseline offers some improvements over using features or the graph structure alone, it never surpasses the strongest signal provider in the graph, never being better than the best one between k-means(features) and SBM.
K-means applied over the extracted DeepWalk representations are also almost never stronger than the community detection using direct SBM likelihood optimization.
}

\begin{table}[!htb]
\footnotesize
\centering{
\newcolumntype{R}{>{\raggedleft\arraybackslash}X}
\newcolumntype{C}{>{\centering\arraybackslash}X}
\newcolumntype{S}{>{\centering\arraybackslash\hsize=.8\hsize}X}
\caption{Results on four datasets from~\cite{shchur2018} in terms of graph conductance $\con$, modularity $\modu$, NMI with ground-truth labels, and pairwise F1 measure. \ant{We group the methods into three categories: baselines using only one aspect of data, neural representation learning, and neural graph pooling methods.} We highlight best neural method performance.}
\begin{tabularx}{\linewidth}{@{}p{1.1cm}SSCSSSCSSSCSSSCS}
\toprule
\multicolumn{1}{C}{} & \multicolumn{4}{c}{\textbf{\damazoncomputers}} & \multicolumn{4}{c}{\textbf{\damazonphoto}} & \multicolumn{4}{c}{\textbf{\dcoauthorcs}} & \multicolumn{4}{c}{\textbf{\dcoauthoreng}} \\
\multicolumn{1}{C}{} & \multicolumn{2}{c}{\emph{graph}} & \multicolumn{2}{c}{\emph{labels}} & \multicolumn{2}{c}{\emph{graph}} & \multicolumn{2}{c}{\emph{labels}} & \multicolumn{2}{c}{\emph{graph}} & \multicolumn{2}{c}{\emph{labels}} & \multicolumn{2}{c}{\emph{graph}} & \multicolumn{2}{c}{\emph{labels}}  \\
\cmidrule(lr){2-5}\cmidrule(lr){6-9}\cmidrule(lr){10-13}\cmidrule(lr){14-17}
\emph{method} & $\con\downarrow$ & $\modu\uparrow$ & NMI$\uparrow$ & F1$\uparrow$ & $\con\downarrow$ & $\modu\uparrow$ & NMI$\uparrow$ & F1$\uparrow$ & $\con\downarrow$ & $\modu\uparrow$ & NMI$\uparrow$ & F1$\uparrow$ & $\con\downarrow$ & $\modu\uparrow$ & NMI$\uparrow$ & F1$\uparrow$\\
\midrule
\mbox{k-m(feat)} & 84.5 & 5.4 & 21.1 & 19.2 & 79.6 & 10.5 & 28.8 & 19.5 & 49.1 & 23.1 & 35.7 & 39.4 & 42.7 & 27.1 & 24.5 & 32.5 \\
\mbox{SBM} & 31.0 & 60.8 & 48.4 & 34.6 & 18.6 & 72.7 & 59.3 & 47.4 & 20.3 & 72.7 & 58.0 & 47.7 & 15.8 & 77.0 & 33.3 & 27.5 \\
\midrule
\mbox{k-m(DW)} & 67.6 & 11.8 & 38.2 & 22.7 & 60.6 & 22.9 & 49.4 & 33.8 & 33.1 & 59.4 &  \textbf{72.7} &  \textbf{61.2} & \textbf{5.7} & 67.4 & 47.7 & 50.0 \\
\mbox{AGC} & 43.2 & 42.8 & 51.3 & 35.3 & 33.8 & 55.9 & 59.0 & 44.2 & 41.5 & 40.1 & 43.3 & 31.9 & 32.3 & 46.4 & 30.8 & 31.2 \\
\mbox{DAEGC} & 39.0 & 43.3 & 42.5 & 37.3 & 19.3 & 58.0 & 47.6 & 45.0 & 39.4 & 49.1 & 36.3 & 32.4 & 31.9 & 50.9 & 12.5 & 26.1 \\
\mbox{SDCN} & 25.1 & 45.6 & 24.9 & 45.2 & 19.7 & 53.3 & 41.7 & 45.1 & 33.0 & 55.7 & 59.3 & 54.7 & 21.8 & 64.6 & 45.3 & 45.9 \\
\mbox{k-m(DGI)} & 61.9 & 22.8 & 22.6 & 15.0 & 51.5 & 35.1 & 33.4 & 23.6 & 35.1 & 57.8 & 64.6 & 51.9 & 29.3 & 60.4 & 49.7 & 37.2 \\
\midrule
\mbox{NOCD} & 26.4 &  \textbf{59.0} & 44.8 & 37.8 & 13.7 &  \textbf{70.1} & 62.3 & 60.2 & 20.9 & 72.2 & 70.5 & 56.4 & 16.0 & \textbf{75.6} & 50.7 & 35.4 \\
\mbox{DiffPool} & 35.6 & 30.4 & 22.1 & 38.3 & 26.5 & 46.8 & 35.9 & 41.8 & 33.6 & 59.3 & 41.6 & 34.4 & 34.9 & 55.0 & 22.0 & 21.8 \\
\mbox{MinCut} & \multicolumn{8}{c}{\textit{did not converge}} & 22.7 & 70.5 & 64.6 & 47.8 & 18.2 & 74.8 & 42.4 & 27.8 \\
\mbox{Ortho} & \multicolumn{8}{c}{\textit{did not converge}} & 27.8 & 65.7 & 64.6 & 46.1 & 18.1 & 74.8 & 43.2 & 28.1 \\
\midrule
\mbox{\thiswork} & \textbf{18.0} & \textbf{59.0} & \textbf{49.3} &  \textbf{45.4} & \textbf{12.7} & \textbf{70.1} & \textbf{63.3} & \textbf{61.0} & \textbf{17.5} & \textbf{72.4} & 69.1 & 59.8 & 7.6 & 72.1 & \textbf{51.8} & \textbf{55.3} \\
\bottomrule
\end{tabularx}}
\label{tbl:clustering2}
\end{table} 
\subsection{Real-world Data}\label{ssec:real-experiments}
We now move on to studies on real-world networks, featuring \thiswork and 7 baselines on 7 different datasets.
\thiswork achieves better clustering performance than its neural counterparts on every single dataset and metric besides losing twice to DGI+k-means on Cora and Citeseer in terms of NMI.
\ant{Compared to SBM, a method that exclusively optimizes modularity, we are able to stay within 3\% in terms of modularity, while simultaneously clustering the features.}
Surprisingly, on Citeseer and Pubmed we achieve better modularity than the method optimizing it directly --- we attribute that to very high correlation between graph structure and the features.

\ant{Compared to other pooling methods, \thiswork improves by over 40\% in conductance, modularity and NMI on average.
In particular, DiffPool achieves overall poor performance across metrics due to its quadratic reconstruction term that does not scale well with graph sparsity (cf.\ Scenario 5).
MinCut pooling performs better, but only manages to match the performance of non-pooling neural representation learning methods on one dataset in terms of ground-truth label NMI.
On \damazoncomputers and \damazonphoto both MinCutPool and its orthogonality-only version failed to converge, even with tuning the parameters.
We attribute that to extremely uneven structure of these graphs, as popular products are co-purchased with a lot of other items, so the effects discussed in~\citet{epasto2017ego,leskovec2008statistical} are prohibiting good cuts.
This corresponds to high values of $d$ ad $d_{\max}$ in our synthetic scenarios 5 and 6.
We also highlight that we beat MinCutPool in terms of conductance (average graph cut) on all datasets, even though it attempts to optimize for this metric.}

\antnew{On large-scale datasets presented in Table~\ref{tbl:clustering-large}, \thiswork achieves 16\% better performance on average across all metrics compared to other end-to-end learning methods.
Some methods, like AGC, DAEGC, and DiffPool are not able to scale to large-scale graphs due to their quadratic time complexity.}

Overall, \thiswork demonstrates excellent performance on both graph clustering and label correlation, successfully leveraging both graph and attribute information. 
Both synthetic and real-world experiments prove that \thiswork is vastly superior to its counterparts in attributed graph clustering.

\begin{table}[!htb]
\footnotesize
\centering{
\newcolumntype{R}{>{\raggedleft\arraybackslash}X}
\newcolumntype{C}{>{\centering\arraybackslash}X}
\newcolumntype{S}{>{\centering\arraybackslash\hsize=.8\hsize}X}
\caption{Results on four large-scale datasets in terms of graph conductance $\con$, modularity $\modu$, NMI with ground-truth labels, and pairwise F1 measure. \ant{We group the methods into three categories: baselines using only one aspect of data, neural representation learning, and neural graph pooling methods.} We highlight best neural method performance.\label{tbl:clustering-large}}
\begin{tabularx}{\linewidth}{@{}p{1.1cm}SSCSSSCSSSCSSSCS}
\toprule
\multicolumn{1}{C}{} & \multicolumn{4}{c}{\textbf{\dcoauthorphy}} & \multicolumn{4}{c}{\textbf{\dcoauthorchem}} & \multicolumn{4}{c}{\textbf{\dcoauthormed}} & \multicolumn{4}{c}{\textbf{\darxiv}} \\
\multicolumn{1}{C}{} & \multicolumn{2}{c}{\emph{graph}} & \multicolumn{2}{c}{\emph{labels}} & \multicolumn{2}{c}{\emph{graph}} & \multicolumn{2}{c}{\emph{labels}} & \multicolumn{2}{c}{\emph{graph}} & \multicolumn{2}{c}{\emph{labels}} & \multicolumn{2}{c}{\emph{graph}} & \multicolumn{2}{c}{\emph{labels}}  \\
\cmidrule(lr){2-5}\cmidrule(lr){6-9}\cmidrule(lr){10-13}\cmidrule(lr){14-17}
\emph{method} & $\con\downarrow$ & $\modu\uparrow$ & NMI$\uparrow$ & F1$\uparrow$ & $\con\downarrow$ & $\modu\uparrow$ & NMI$\uparrow$ & F1$\uparrow$ & $\con\downarrow$ & $\modu\uparrow$ & NMI$\uparrow$ & F1$\uparrow$ & $\con\downarrow$ & $\modu\uparrow$ & NMI$\uparrow$ & F1$\uparrow$\\
\midrule
\mbox{k-m(feat)} & 57.0 & 19.4 & 30.6 & 42.9 & 42.9 & 18.2 & 13.9 & 35.1 & 54.7 & 19.3 & 11.8 & 31.7 & 75.9 & 16.4 & 20.3 & 20.2 \\
\mbox{SBM} & 25.9 & 66.9 & 45.4 & 30.4 & 18.4 & 74.6 & 25.4 & 25.0 & 21.1 & 72.0 & 36.1 & 31.1 & 24.8 & 67.6 & 31.9 & 28.3 \\
\midrule
\mbox{k-m(DW)} & 44.7 & 47.0 & 43.5 & 24.3 & \textbf{14.0} & \textbf{74.8} & 36.5 & 33.8 & \textbf{16.6} & \textbf{72.1} & 43.1 & 39.4 & 29.5 & \textbf{58.2} & 28.4 & 36.0 \\
\mbox{SDCN} & 32.1 & 52.8 & 50.4 & 39.9 & 29.9 & 58.7 & 33.3 & 32.8 & 34.8 & 54.2 & 25.2 & 26.5 & 31.2 & 36.8 & 15.3 & 28.8 \\
\mbox{k-m(DGI)} & 38.6 & 51.2 & 51.0 & 30.6 & 31.6 & 60.6 & 40.8 & 32.9 & 35.7 & 56.5 & 34.8 & 27.7 & 58.7 & 29.7 & 30.0 & 24.6 \\
\midrule
\mbox{NOCD} & 25.7 & 65.5 &  51.9 & 28.7 & 19.2 & 73.1 & \textbf{43.1} & 40.1 & 22.0 & 69.7 & 42.5 & 37.6 & 41.1 & 41.9 & 20.7 & 38.2 \\
\mbox{MinCut} & 27.8 & 64.3 & 48.3 & 24.9 & 21.2 & 72.2 & 39.0 & 32.0 & 22.8 & 69.7 & 40.0 & 32.1 & 37.4 & 52.6 & 36.0 & 27.1 \\
\mbox{Ortho} & 33.0 & 59.5 & 44.7 & 23.7 & 21.6 & 71.8 & 38.5 & 31.0 & 22.9 & 69.5 & 40.4 & 32.1 & 37.8 & 52.2 & 35.6 & 26.7 \\
\midrule
\mbox{\thiswork}& \textbf{18.8} & \textbf{65.8} & \textbf{56.7} & \textbf{42.4}  & 16.2 & 73.9 & \textbf{43.2} & \textbf{43.7} & 17.6 & 71.6 & \textbf{43.4} & \textbf{40.0} & \textbf{22.3} & {57.4} & \textbf{37.6} & \textbf{45.7} \\
\bottomrule
\end{tabularx}}
\end{table}  \section{Conclusion}\label{sec:conclusion}
In this work, we study GNN pooling through the lens of attributed graph clustering.
We introduce Deep Modularity Networks (\thiswork), an unsupervised objective and realize it with a GNN which can recover high quality clusters.
We compare against challenging baselines that baselines that optimize structure (SBM), features (kmeans), or both (DGI+k-means), in addition to a recently proposed state-of-the-art pooling method (MinCutPool).

We explore the limits of GNN clustering methods in terms of both graph and feature signals using synthetic data, where we see that \thiswork better leverages structure and attributes than all existing methods.
In extensive experiments on real datasets we show that the clusters found by \thiswork are more likely to correspond to ground truth labels, and have better properties as illustrated by clustering metrics (e.g.\ conductance or modularity).
We hope that this work will further advancements in unsupervised learning for GNNs as well as attributed graph clustering, allowing further advances in graph learning. \bibliography{main.bib}

\begin{thebibliography}{77}
\providecommand{\natexlab}[1]{#1}
\providecommand{\url}[1]{\texttt{#1}}
\expandafter\ifx\csname urlstyle\endcsname\relax
  \providecommand{\doi}[1]{doi: #1}\else
  \providecommand{\doi}{doi: \begingroup \urlstyle{rm}\Url}\fi

\bibitem[Abbe and Sandon(2015)]{abbe2015community}
Emmanuel Abbe and Colin Sandon.
\newblock Community detection in general stochastic block models: Fundamental
  limits and efficient algorithms for recovery.
\newblock In \emph{FOCS}. IEEE, 2015.

\bibitem[Arthur and Vassilvitskii(2007)]{arthur2007}
David Arthur and Sergei Vassilvitskii.
\newblock {K}-means++: The advantages of careful seeding.
\newblock In \emph{SODA}, 2007.

\bibitem[Bansal et~al.(2018)Bansal, Chen, and Wang]{bansal2018can}
Nitin Bansal, Xiaohan Chen, and Zhangyang Wang.
\newblock Can we gain more from orthogonality regularizations in training deep
  {CNNs}?
\newblock In \emph{NIPS}, 2018.

\bibitem[Belghazi et~al.(2018)Belghazi, Baratin, Rajeswar, Ozair, Bengio,
  Courville, and Hjelm]{belghazi2018}
Mohamed~Ishmael Belghazi, Aristide Baratin, Sai Rajeswar, Sherjil Ozair, Yoshua
  Bengio, Aaron Courville, and R~Devon Hjelm.
\newblock Mine: mutual information neural estimation.
\newblock In \emph{ICML}, 2018.

\bibitem[Bianchi et~al.(2020)Bianchi, Grattarola, and Alippi]{bianchi2020}
Filippo~Maria Bianchi, Daniele Grattarola, and Cesare Alippi.
\newblock Spectral clustering with graph neural networks for graph pooling.
\newblock In \emph{ICML}, 2020.

\bibitem[Bickel and Chen(2009)]{bickel2009nonparametric}
Peter~J Bickel and Aiyou Chen.
\newblock A nonparametric view of network models and newman--girvan and other
  modularities.
\newblock \emph{Proceedings of the National Academy of Sciences}, 2009.

\bibitem[Bo et~al.(2020)Bo, Wang, Shi, Zhu, Lu, and Cui]{bo2020structural}
Deyu Bo, Xiao Wang, Chuan Shi, Meiqi Zhu, Emiao Lu, and Peng Cui.
\newblock Structural deep clustering network.
\newblock In \emph{Proceedings of The Web Conference 2020}, pages 1400--1410,
  2020.

\bibitem[Brandes et~al.(2006)Brandes, Delling, Gaertler, G{\"o}rke, Hoefer,
  Nikoloski, and Wagner]{brandes2006maximizing}
Ulrik Brandes, Daniel Delling, Marco Gaertler, Robert G{\"o}rke, Martin Hoefer,
  Zoran Nikoloski, and Dorothea Wagner.
\newblock Maximizing modularity is hard.
\newblock \emph{arXiv preprint physics/0608255}, 2006.

\bibitem[Bronstein et~al.(2017)Bronstein, Bruna, LeCun, Szlam, and
  Vandergheynst]{bronstein2017}
Michael~M Bronstein, Joan Bruna, Yann LeCun, Arthur Szlam, and Pierre
  Vandergheynst.
\newblock Geometric deep learning: going beyond euclidean data.
\newblock \emph{IEEE Signal Processing Magazine}, 2017.

\bibitem[Cabreros et~al.(2016)Cabreros, Abbe, and
  Tsirigos]{cabreros2016detecting}
Irineo Cabreros, Emmanuel Abbe, and Aristotelis Tsirigos.
\newblock Detecting community structures in hi-c genomic data.
\newblock In \emph{2016 Annual Conference on Information Science and Systems
  (CISS)}, pages 584--589. IEEE, 2016.

\bibitem[Chami et~al.(2022)Chami, Abu-El-Haija, Perozzi, R{\'e}, and
  Murphy]{chami2020}
Ines Chami, Sami Abu-El-Haija, Bryan Perozzi, Christopher R{\'e}, and Kevin
  Murphy.
\newblock Machine learning on graphs: A model and comprehensive taxonomy.
\newblock \emph{JMLR}, 2022.

\bibitem[Chen et~al.(2018)Chen, Perozzi, Hu, and Skiena]{chen2018}
Haochen Chen, Bryan Perozzi, Yifan Hu, and Steven Skiena.
\newblock {HARP}: Hierarchical representation learning for networks.
\newblock In \emph{AAAI}, 2018.

\bibitem[Cl{\'e}men{\c{c}}on et~al.(2012)Cl{\'e}men{\c{c}}on, De~Arazoza,
  Rossi, and Tran]{clemenccon2012hierarchical}
St{\'e}phan Cl{\'e}men{\c{c}}on, Hector De~Arazoza, Fabrice Rossi, and Viet~Chi
  Tran.
\newblock Hierarchical clustering for graph visualization.
\newblock \emph{arXiv preprint arXiv:1210.5693}, 2012.

\bibitem[Cui et~al.(2008)Cui, Zhou, Qu, Wong, and Li]{cui2008geometry}
Weiwei Cui, Hong Zhou, Huamin Qu, Pak~Chung Wong, and Xiaoming Li.
\newblock Geometry-based edge clustering for graph visualization.
\newblock \emph{IEEE Transactions on Visualization and Computer Graphics},
  14\penalty0 (6):\penalty0 1277--1284, 2008.

\bibitem[Defferrard et~al.(2016)Defferrard, Bresson, and
  Vandergheynst]{defferrard2016}
Micha{\"e}l Defferrard, Xavier Bresson, and Pierre Vandergheynst.
\newblock Convolutional neural networks on graphs with fast localized spectral
  filtering.
\newblock In \emph{NIPS}, 2016.

\bibitem[Deng et~al.(2020)Deng, Zhao, Wang, Zhang, and Feng]{deng2020}
Chenhui Deng, Zhiqiang Zhao, Yongyu Wang, Zhiru Zhang, and Zhuo Feng.
\newblock {GraphZoom}: A multi-level spectral approach for accurate and
  scalable graph embedding.
\newblock In \emph{ICLR}, 2020.

\bibitem[Duvenaud et~al.(2015)Duvenaud, Maclaurin, Iparraguirre, Bombarell,
  Hirzel, Aspuru-Guzik, and Adams]{duvenaud2015}
David~K Duvenaud, Dougal Maclaurin, Jorge Iparraguirre, Rafael Bombarell,
  Timothy Hirzel, Al{\'a}n Aspuru-Guzik, and Ryan~P Adams.
\newblock Convolutional networks on graphs for learning molecular fingerprints.
\newblock In \emph{NIPS}, 2015.

\bibitem[Dwivedi et~al.(2023)Dwivedi, Joshi, Luu, Laurent, Bengio, and
  Bresson]{dwivedi2020benchmarking}
Vijay~Prakash Dwivedi, Chaitanya~K. Joshi, Anh~Tuan Luu, Thomas Laurent, Yoshua
  Bengio, and Xavier Bresson.
\newblock Benchmarking graph neural networks.
\newblock \emph{Journal of Machine Learning Research}, 24\penalty0
  (43):\penalty0 1--48, 2023.
\newblock URL \url{http://jmlr.org/papers/v24/22-0567.html}.

\bibitem[Epasto et~al.(2017)Epasto, Lattanzi, and Paes~Leme]{epasto2017ego}
Alessandro Epasto, Silvio Lattanzi, and Renato Paes~Leme.
\newblock Ego-splitting framework: From non-overlapping to overlapping
  clusters.
\newblock In \emph{KDD}, 2017.

\bibitem[Fiedler(1973)]{fiedler1973algebraic}
Miroslav Fiedler.
\newblock Algebraic connectivity of graphs.
\newblock \emph{Czechoslovak mathematical journal}, 1973.

\bibitem[Fortunato and Hric(2016)]{fortunato2016community}
Santo Fortunato and Darko Hric.
\newblock Community detection in networks: A user guide.
\newblock \emph{Physics reports}, 2016.

\bibitem[Gao and Ji(2019)]{gao2019}
Hongyang Gao and Shuiwang Ji.
\newblock Graph {U}-nets.
\newblock In \emph{ICML}, 2019.

\bibitem[Gilmer et~al.(2017)Gilmer, Schoenholz, Riley, Vinyals, and
  Dahl]{gilmer2017}
Justin Gilmer, Samuel~S Schoenholz, Patrick~F Riley, Oriol Vinyals, and
  George~E Dahl.
\newblock Neural message passing for quantum chemistry.
\newblock In \emph{ICML}, 2017.

\bibitem[Good et~al.(2010)Good, De~Montjoye, and Clauset]{good2010performance}
Benjamin~H Good, Yves-Alexandre De~Montjoye, and Aaron Clauset.
\newblock Performance of modularity maximization in practical contexts.
\newblock \emph{Physical Review E}, 81\penalty0 (4):\penalty0 046106, 2010.

\bibitem[Grover and Leskovec(2016)]{grover2016}
Aditya Grover and Jure Leskovec.
\newblock node2vec: Scalable feature learning for networks.
\newblock In \emph{KDD}, 2016.

\bibitem[Gutmann and Hyv{\"a}rinen(2010)]{gutmann2010}
Michael Gutmann and Aapo Hyv{\"a}rinen.
\newblock Noise-contrastive estimation: A new estimation principle for
  unnormalized statistical models.
\newblock In \emph{AISTATS}, 2010.

\bibitem[Hamilton et~al.(2017)Hamilton, Ying, and Leskovec]{hamilton2017survey}
William~L Hamilton, Rex Ying, and Jure Leskovec.
\newblock Representation learning on graphs: Methods and applications.
\newblock \emph{IEEE Data Engineering Bulletin}, 2017.

\bibitem[Hjelm et~al.(2019)Hjelm, Fedorov, Lavoie-Marchildon, Grewal, Bachman,
  Trischler, and Bengio]{hjelm2018}
R~Devon Hjelm, Alex Fedorov, Samuel Lavoie-Marchildon, Karan Grewal, Phil
  Bachman, Adam Trischler, and Yoshua Bengio.
\newblock Learning deep representations by mutual information estimation and
  maximization.
\newblock In \emph{ICLR}, 2019.

\bibitem[Hu et~al.(2020)Hu, Fey, Zitnik, Dong, Ren, Liu, Catasta, and
  Leskovec]{hu2020open}
Weihua Hu, Matthias Fey, Marinka Zitnik, Yuxiao Dong, Hongyu Ren, Bowen Liu,
  Michele Catasta, and Jure Leskovec.
\newblock Open graph benchmark: Datasets for machine learning on graphs.
\newblock \emph{NeurIPS}, 2020.

\bibitem[Hu* et~al.(2020)Hu*, Liu*, Gomes, Zitnik, Liang, Pande, and
  Leskovec]{hu2020}
Weihua Hu*, Bowen Liu*, Joseph Gomes, Marinka Zitnik, Percy Liang, Vijay Pande,
  and Jure Leskovec.
\newblock Strategies for pre-training graph neural networks.
\newblock In \emph{ICLR}, 2020.

\bibitem[Karrer and Newman(2011)]{karrer2011stochastic}
Brian Karrer and Mark~EJ Newman.
\newblock Stochastic blockmodels and community structure in networks.
\newblock \emph{Physical review E}, 83\penalty0 (1):\penalty0 016107, 2011.

\bibitem[Kernighan and Lin(1970)]{kernighan1970efficient}
Brian~W Kernighan and Shen Lin.
\newblock An efficient heuristic procedure for partitioning graphs.
\newblock \emph{The Bell system technical journal}, 1970.

\bibitem[Kipf and Welling(2017)]{kipf2017}
Thomas~N Kipf and Max Welling.
\newblock Semi-supervised classification with graph convolutional networks.
\newblock In \emph{ICLR}, 2017.

\bibitem[Klambauer et~al.(2017)Klambauer, Unterthiner, Mayr, and
  Hochreiter]{klambauer2017self}
G{\"u}nter Klambauer, Thomas Unterthiner, Andreas Mayr, and Sepp Hochreiter.
\newblock Self-normalizing neural networks.
\newblock In \emph{NIPS}, 2017.

\bibitem[Lee et~al.(2019)Lee, Lee, and Kang]{lee2019}
Junhyun Lee, Inyeop Lee, and Jaewoo Kang.
\newblock Self-attention graph pooling.
\newblock In \emph{ICML}, 2019.

\bibitem[Leskovec et~al.(2008)Leskovec, Lang, Dasgupta, and
  Mahoney]{leskovec2008statistical}
Jure Leskovec, Kevin~J Lang, Anirban Dasgupta, and Michael~W Mahoney.
\newblock Statistical properties of community structure in large social and
  information networks.
\newblock In \emph{WWW}, 2008.

\bibitem[Liang et~al.(2021)Liang, Gurukar, and Parthasarathy]{liang2018}
Jiongqian Liang, Saket Gurukar, and Srinivasan Parthasarathy.
\newblock {MILE}: A multi-level framework for scalable graph embedding.
\newblock In \emph{ICWSM}, 2021.

\bibitem[Lloyd(1982)]{lloyd1982}
Stuart Lloyd.
\newblock Least squares quantization in {PCM}.
\newblock \emph{IEEE Transactions on Information Theory}, 1982.

\bibitem[McDiarmid and Skerman(2020)]{mcdiarmid2020modularity}
Colin McDiarmid and Fiona Skerman.
\newblock Modularity of erd{\H{o}}s-r{\'e}nyi random graphs.
\newblock \emph{Random Structures \& Algorithms}, 2020.

\bibitem[Mothe et~al.(2017)Mothe, Mkhitaryan, and
  Haroutunian]{mothe2017community}
Josiane Mothe, Karen Mkhitaryan, and Mariam Haroutunian.
\newblock Community detection: Comparison of state of the art algorithms.
\newblock In \emph{CSIT}. IEEE, 2017.

\bibitem[Nadakuditi and Newman(2012)]{nadakuditi2012graph}
Raj~Rao Nadakuditi and Mark~EJ Newman.
\newblock Graph spectra and the detectability of community structure in
  networks.
\newblock \emph{Physical review letters}, 2012.

\bibitem[Navarin et~al.(2018)Navarin, Tran, and Sperduti]{navarin2018}
Nicol{\`o} Navarin, Dinh~V Tran, and Alessandro Sperduti.
\newblock Pre-training graph neural networks with kernels.
\newblock In \emph{NeurIPS}, 2018.

\bibitem[Newman(2006{\natexlab{a}})]{newman2006}
Mark~EJ Newman.
\newblock Modularity and community structure in networks.
\newblock \emph{PNAS}, 2006{\natexlab{a}}.

\bibitem[Newman(2006{\natexlab{b}})]{newman2006finding}
Mark~EJ Newman.
\newblock Finding community structure in networks using the eigenvectors of
  matrices.
\newblock \emph{Physical review E}, 2006{\natexlab{b}}.

\bibitem[Niepert et~al.(2016)Niepert, Ahmed, and Kutzkov]{niepert2016}
Mathias Niepert, Mohamed Ahmed, and Konstantin Kutzkov.
\newblock Learning convolutional neural networks for graphs.
\newblock In \emph{ICML}, 2016.

\bibitem[Nowicki and Snijders(2001)]{nowicki2001estimation}
Krzysztof Nowicki and Tom A~B Snijders.
\newblock Estimation and prediction for stochastic blockstructures.
\newblock \emph{Journal of the American statistical association}, 96\penalty0
  (455):\penalty0 1077--1087, 2001.

\bibitem[Papacharissi(2009)]{papacharissi2009virtual}
Zizi Papacharissi.
\newblock The virtual geographies of social networks: a comparative analysis of
  facebook, linkedin and asmallworld.
\newblock \emph{New media \& society}, 2009.

\bibitem[Peixoto(2014{\natexlab{a}})]{peixoto2014}
Tiago~P Peixoto.
\newblock Efficient monte carlo and greedy heuristic for the inference of
  stochastic block models.
\newblock \emph{Physical Review E}, 2014{\natexlab{a}}.

\bibitem[Peixoto(2014{\natexlab{b}})]{peixoto_graph-tool_2014}
Tiago~P. Peixoto.
\newblock The graph-tool python library.
\newblock \emph{figshare}, 2014{\natexlab{b}}.
\newblock \doi{10.6084/m9.figshare.1164194}.
\newblock URL \url{http://figshare.com/articles/graph_tool/1164194}.

\bibitem[Perozzi and Akoglu(2016)]{perozzi2016scalable}
Bryan Perozzi and Leman Akoglu.
\newblock Scalable anomaly ranking of attributed neighborhoods.
\newblock In \emph{SDM}, 2016.

\bibitem[Perozzi and Akoglu(2018)]{perozzi2018discovering}
Bryan Perozzi and Leman Akoglu.
\newblock Discovering communities and anomalies in attributed graphs:
  Interactive visual exploration and summarization.
\newblock \emph{ACM TKDE}, 12\penalty0 (2), 2018.

\bibitem[Perozzi et~al.(2014{\natexlab{a}})Perozzi, Akoglu,
  Iglesias~S\'{a}nchez, and M\"{u}ller]{perozzi2014focused}
Bryan Perozzi, Leman Akoglu, Patricia Iglesias~S\'{a}nchez, and Emmanuel
  M\"{u}ller.
\newblock Focused clustering and outlier detection in large attributed graphs.
\newblock In \emph{KDD}, 2014{\natexlab{a}}.

\bibitem[Perozzi et~al.(2014{\natexlab{b}})Perozzi, Al-Rfou, and
  Skiena]{perozzi2014}
Bryan Perozzi, Rami Al-Rfou, and Steven Skiena.
\newblock Deep{W}alk: Online learning of social representations.
\newblock In \emph{KDD}, 2014{\natexlab{b}}.

\bibitem[Scarselli et~al.(2008)Scarselli, Gori, Tsoi, Hagenbuchner, and
  Monfardini]{scarselli2008}
Franco Scarselli, Marco Gori, Ah~Chung Tsoi, Markus Hagenbuchner, and Gabriele
  Monfardini.
\newblock The graph neural network model.
\newblock \emph{IEEE Transactions on Neural Networks}, 2008.

\bibitem[Sen et~al.(2008)Sen, Namata, Bilgic, Getoor, Galligher, and
  Eliassi-Rad]{sen2008}
Prithviraj Sen, Galileo Namata, Mustafa Bilgic, Lise Getoor, Brian Galligher,
  and Tina Eliassi-Rad.
\newblock Collective classification in network data.
\newblock \emph{AI magazine}, 2008.

\bibitem[Shchur and G{\"u}nnemann(2019)]{shchur2019overlapping}
Oleksandr Shchur and Stephan G{\"u}nnemann.
\newblock Overlapping community detection with graph neural networks.
\newblock \emph{arXiv preprint arXiv:1909.12201}, 2019.

\bibitem[Shchur et~al.(2018)Shchur, Mumme, Bojchevski, and
  G{\"u}nnemann]{shchur2018}
Oleksandr Shchur, Maximilian Mumme, Aleksandar Bojchevski, and Stephan
  G{\"u}nnemann.
\newblock Pitfalls of graph neural network evaluation.
\newblock \emph{arXiv preprint arXiv:1811.05868}, 2018.

\bibitem[Shi and Malik(2000)]{shi2000normalized}
Jianbo Shi and Jitendra Malik.
\newblock Normalized cuts and image segmentation.
\newblock \emph{IEEE Transactions on pattern analysis and machine
  intelligence}, 2000.

\bibitem[Snijders and Nowicki(1997)]{snijders1997estimation}
Tom~AB Snijders and Krzysztof Nowicki.
\newblock Estimation and prediction for stochastic blockmodels for graphs with
  latent block structure.
\newblock \emph{Journal of classification}, 1997.

\bibitem[Song and Ermon(2020)]{song2019}
Jiaming Song and Stefano Ermon.
\newblock Understanding the limitations of variational mutual information
  estimators.
\newblock In \emph{ICLR}, 2020.

\bibitem[Srivastava et~al.(2014)Srivastava, Hinton, Krizhevsky, Sutskever, and
  Salakhutdinov]{srivastava2014dropout}
Nitish Srivastava, Geoffrey Hinton, Alex Krizhevsky, Ilya Sutskever, and Ruslan
  Salakhutdinov.
\newblock Dropout: a simple way to prevent neural networks from overfitting.
\newblock \emph{JMLR}, 2014.

\bibitem[Sun et~al.(2020)Sun, Hoffmann, and Tang]{sun2020}
Fan-Yun Sun, Jordan Hoffmann, and Jian Tang.
\newblock Info{G}raph: Unsupervised and semi-supervised graph-level
  representation learning via mutual information maximization.
\newblock In \emph{ICLR}, 2020.

\bibitem[Traud et~al.(2011)Traud, Kelsic, Mucha, and
  Porter]{traud2011comparing}
Amanda~L Traud, Eric~D Kelsic, Peter~J Mucha, and Mason~A Porter.
\newblock Comparing community structure to characteristics in online collegiate
  social networks.
\newblock \emph{SIAM review}, 2011.

\bibitem[Tschannen et~al.(2020)Tschannen, Djolonga, Rubenstein, Gelly, and
  Lucic]{tschannen2019}
Michael Tschannen, Josip Djolonga, Paul~K Rubenstein, Sylvain Gelly, and Mario
  Lucic.
\newblock On mutual information maximization for representation learning.
\newblock In \emph{ICLR}, 2020.

\bibitem[Tsitsulin et~al.(2018)Tsitsulin, Mottin, Karras, and
  M{\"u}ller]{tsitsulin2018}
Anton Tsitsulin, Davide Mottin, Panagiotis Karras, and Emmanuel M{\"u}ller.
\newblock {VERSE}: Versatile graph embeddings from similarity measures.
\newblock In \emph{WWW}, 2018.

\bibitem[Veli{\v{c}}kovi{\'c} et~al.(2017)Veli{\v{c}}kovi{\'c}, Cucurull,
  Casanova, Romero, Li{\`o}, and Bengio]{velickovic2017}
Petar Veli{\v{c}}kovi{\'c}, Guillem Cucurull, Arantxa Casanova, Adriana Romero,
  Pietro Li{\`o}, and Yoshua Bengio.
\newblock Graph attention networks.
\newblock In \emph{ICLR}, 2017.

\bibitem[Veli{\v{c}}kovi{\'c} et~al.(2019)Veli{\v{c}}kovi{\'c}, Fedus,
  Hamilton, Li{\`o}, Bengio, and Hjelm]{velickovic2018}
Petar Veli{\v{c}}kovi{\'c}, William Fedus, William~L Hamilton, Pietro Li{\`o},
  Yoshua Bengio, and R~Devon Hjelm.
\newblock Deep graph infomax.
\newblock In \emph{ICLR}, 2019.

\bibitem[Wang et~al.(2019{\natexlab{a}})Wang, Pan, Hu, Long, Jiang, and
  Zhang]{wang2019attributed}
Chun Wang, Shirui Pan, Ruiqi Hu, Guodong Long, Jing Jiang, and Chengqi Zhang.
\newblock Attributed graph clustering: A deep attentional embedding approach.
\newblock In \emph{IJCAI}, 2019{\natexlab{a}}.

\bibitem[Wang et~al.(2019{\natexlab{b}})Wang, Zheng, Li, and
  Wang]{wang2019cvpr}
Zhongdao Wang, Liang Zheng, Yali Li, and Shengjin Wang.
\newblock Linkage based face clustering via graph convolution network.
\newblock In \emph{CVPR}, 2019{\natexlab{b}}.

\bibitem[Wei and Cheng(1989)]{wei1989towards}
Yen-Chuen Wei and Chung-Kuan Cheng.
\newblock Towards efficient hierarchical designs by ratio cut partitioning.
\newblock In \emph{IEEE International Conference on Computer-Aided Design}.
  IEEE, 1989.

\bibitem[Yang and Leskovec(2015)]{yang2015}
Jaewon Yang and Jure Leskovec.
\newblock Defining and evaluating network communities based on ground-truth.
\newblock \emph{Knowledge and Information Systems}, 2015.

\bibitem[Yang et~al.(2019)Yang, Zhan, Chen, Yan, Loy, and Lin]{yang2019}
Lei Yang, Xiaohang Zhan, Dapeng Chen, Junjie Yan, Chen~Change Loy, and Dahua
  Lin.
\newblock Learning to cluster faces on an affinity graph.
\newblock In \emph{CVPR}, 2019.

\bibitem[Ying et~al.(2018{\natexlab{a}})Ying, He, Chen, Eksombatchai, Hamilton,
  and Leskovec]{ying2018graph}
Rex Ying, Ruining He, Kaifeng Chen, Pong Eksombatchai, William~L Hamilton, and
  Jure Leskovec.
\newblock Graph convolutional neural networks for web-scale recommender
  systems.
\newblock In \emph{KDD}, 2018{\natexlab{a}}.

\bibitem[Ying et~al.(2018{\natexlab{b}})Ying, You, Morris, Ren, Hamilton, and
  Leskovec]{ying2018}
Zhitao Ying, Jiaxuan You, Christopher Morris, Xiang Ren, Will Hamilton, and
  Jure Leskovec.
\newblock Hierarchical graph representation learning with differentiable
  pooling.
\newblock In \emph{NeurIPS}, 2018{\natexlab{b}}.

\bibitem[You et~al.(2019)You, Ying, and Leskovec]{you2019}
Jiaxuan You, Rex Ying, and Jure Leskovec.
\newblock Position-aware graph neural networks.
\newblock In \emph{ICML}, 2019.

\bibitem[Zhang et~al.(2019)Zhang, Liu, Li, and Wu]{zhang2019attributed}
Xiaotong Zhang, Han Liu, Qimai Li, and Xiao-Ming Wu.
\newblock Attributed graph clustering via adaptive graph convolution.
\newblock In \emph{IJCAI}, 2019.

\bibitem[Zhao et~al.(2012)Zhao, Levina, and Zhu]{zhao2012consistency}
Yunpeng Zhao, Elizaveta Levina, and Ji~Zhu.
\newblock Consistency of community detection in networks under degree-corrected
  stochastic block models.
\newblock \emph{The Annals of Statistics}, 2012.

\end{thebibliography}

\end{document}